\documentclass[twoside]{article}

\usepackage[accepted]{aistats2025}
\usepackage{natbib}
\setcitestyle{authoryear,open={(},close={)}}
\usepackage{hyperref}
\usepackage{booktabs}
\usepackage{amsmath}
\usepackage{amsfonts}
\usepackage{amsthm}
\usepackage{amssymb}
\usepackage{algorithm}
\usepackage{algpseudocode}
\usepackage{xcolor}
\usepackage{graphicx}
\usepackage{subcaption}
\usepackage{microtype}
\usepackage{array}
\usepackage{mathtools}
\usepackage[capitalise, noabbrev]{cleveref}

\newcommand{\R}{\mathbb{R}}
\newcommand{\U}{\mathcal{U}}
\newcommand{\bX}{\mathbf{X}}
\newcommand{\Xs}{X^{\star}}
\newcommand{\bY}{\mathbf{Y}}
\newcommand{\Ys}{Y^{\star}}
\newcommand{\Ytheta}{Y^{(\theta)}}
\newcommand{\bx}[1]{\mathbf{x}^{(#1)}}
\newcommand{\hbx}[1]{\hat{\mathbf{x}}^{(#1)}}
\newcommand{\bxs}[1]{\mathbf{x}^{\star(#1)}}
\newcommand{\hbxs}[1]{\hat{\mathbf{x}}^{\star(#1)}}
\newcommand{\bz}[1]{\mathbf{z}^{(#1)}}
\newcommand{\bzs}[1]{\mathbf{z}^{\star(#1)}}
\newcommand{\Ps}{\mathbb{P}^{\star}}
\newcommand{\Pt}{\mathbb{P}^{(\theta)}}
\renewcommand{\L}{\mathcal{L}}

\renewcommand{\d}{\mathrm{d}}
\renewcommand{\H}{\mathcal{H}}
\renewcommand{\P}{\mathbb{P}}
\newcommand{\E}{\mathbb{E}}
\newcommand{\kld}{\operatorname{D}_{\text{KL}}}

\newtheorem{theorem}{Theorem}
\newtheorem{lemma}{Lemma}
\begin{document}

%

%

\twocolumn[

\aistatstitle{Infinite-dimensional Diffusion Bridge Simulation via Operator Learning}



\aistatsauthor{Gefan Yang\(^{1}\) \And Elizabeth L. Baker\(^{1}\) $\qquad$ \And Michael L. Severinsen\(^{2}\) $\quad$ \And Christy A. Hipsley\(^{3}\) \And Stefan Sommer\(^{1}\) }

\aistatsaddress{\(^{1}\)DIKU, UCPH \And \(^{2}\)Globe Institute, UCPH \And \(^{3}\)Department of Biology, UCPH} ]

\begin{abstract}
  The diffusion bridge, which is a diffusion process conditioned on hitting a specific state within a finite period, has found broad applications in various scientific and engineering fields. However, simulating diffusion bridges for modeling natural data can be challenging due to both the intractability of the drift term and continuous representations of the data. Although several methods are available to simulate finite-dimensional diffusion bridges, infinite-dimensional cases remain under explored. This paper presents a method that merges score matching techniques with operator learning, enabling a direct approach to learn the infinite-dimensional bridge and achieving a discretization equivariant bridge simulation. We conduct a series of experiments, ranging from synthetic examples with closed-form solutions to the stochastic nonlinear evolution of real-world biological shape data. Our method demonstrates high efficacy, particularly due to its ability to adapt to any resolution without extra training.
\end{abstract}

\section{INTRODUCTION}
Diffusion processes are commonly utilized in diverse scientific fields, including mathematics, physics, evolutionary biology, and finance, to model stochastic dynamics. Updating the posterior of the model through conditioning on observed data is a crucial element of the process, and there are several methods to achieve this. For example, Gaussian process regression can be used for finite-dimensional or infinite-dimensional linear models \citep{shi2011gaussian}, and finite-dimensional Doob's \(h\)-transform is designed for finite nonlinear processes \citep[Chapter 6]{rogers2000diffusions}. 
On the other hand, many data types, such as images and sound signals, are naturally continuous and described by functions, thereby being infinite-dimensional. Moreover, the stochasticity is also nonlinear. One approach is to discretize them and represent them by vectors using finite-dimensional models. However, a more natural way is to work directly in the infinite-dimensional function space, performing only finite-dimensional projections during inference. This functional formulation offers resolution-invariant features, which are more memory-efficient and demonstrate greater generalization capacities. Simulating infinite-dimensional linear processes has been explored in several studies \citep{franzese2024continuous, lim2023score, pidstrigach2023infinite}. However, extending these methods to condition nonlinear processes has received relatively little exploration except \cite{baker2024conditioning}. 

Interest in the study of linear diffusion processes has surged recently, largely driven by advancements in diffusion generative models.
In diffusion generative models, the data undergoes perturbation via an unconditional linear diffusion process, resulting in noise that follows a simple distribution and is easy to sample. 
The sampled noise is then transformed back to the clean data through a reverse process, which is also linear and can be learned by the score matching techniques \citep{hyvarinen2005estimation, vincent2011connection}.
Recently, the study of diffusion models using stochastic differential equations (SDEs) and their mathematical interpretations has gained significant attention \citep{song2020score, huang2021variational}. Based on the rich research on the time-reversed diffusion processes and the design of neural network approximation, we show that similar techniques can be applied to simulating conditioned nonlinear infinite-dimensional diffusion processes.

Extending the finite-dimensional bridge simulation schemes into infinite dimensions is nontrivial, where the problems that need to be addressed are: 1) the theoretical establishment of the conditioning mechanism, 2) the theoretical soundness of the infinite-dimensional time-reversed nonlinear diffusion bridge, and 3) an efficient and straightforward approach to learning the infinite-dimensional bridge. The first problem has been solved in \cite{baker2024conditioning}, through the \emph{infinite-dimensional Doob's \(h\)-transform}. In this paper, we focus on the remaining two. We first derive the form of the time reversal of the infinite-dimensional diffusion bridge, which lifts the framework in \cite{heng2021simulating} to infinite dimensions. Furthermore, we propose a tractable optimization objective for learning the reverse bridge. We also design a suitable neural operator structure that can learn the bridge efficiently.

\section{RELATED WORK AND CONTRIBUTIONS}
\subsection{Related work}
\textbf{Bridge simulations:} The linear diffusion process always has a Gaussian transition density, which gives the additional drift term in the corresponding conditional process a closed form. The challenge lies in the nonlinear case, where the transition is no longer Gaussian. A mainstream approach is \emph{guided proposals}, originally conceived in \cite{clark1990simulation} and developed further by \cite{delyon2006simulation, schauer2017guided,mider2021continuous}. The essential idea of a guided proposal is to use an approximated bridge with known transition density as the proposal. Then a Markov Chain Monte Carlo (MCMC) algorithm is used to correct the sampling by computing the likelihood ratio between the proposed and true bridges in closed form. However, such a method suffers from expensive covariance matrix inversion in high-dimensional cases, time-costly MCMC updating iterations, and redundancy for adjusting different levels of resolution. Another approach is to directly approximate the unknown drift in the true bridge using either score-based-learning \citep{heng2021simulating, baker2024conditioning} or kernel approximation \citep{chau2024efficient}. \cite{heng2021simulating} used denoising score matching to learn the score from a variational perspective for finite-dimensional diffusion bridges;
\cite{baker2024conditioning} derived an infinite-dimensional Doob's \(h\)-transform for conditioning infinite-dimensional nonlinear processes, and trained a neural network to learn the finite-dimensional base-projection of the infinite-dimensional bridge, extending the score-based bridge simulation scheme to infinite dimensions. \cite{chau2024efficient}, on the other hand, proposed to use Gaussian kernels to approximate the intractable score for finite-dimensional bridges.

\textbf{Infinite-dimensional diffusion models:} (Score-based) diffusion generative models (SGMs) are initially developed to generate samples in Euclidean spaces \citep{sohl2015deep, song2021scorebased}. Various studies have been done to generalize SGMs to infinite-dimensional spaces with either Hilbert-space-defined score functions \citep{pidstrigach2023infinite, baldassari2024conditional, lim2023score} or finite-dimensional score projections \citep{franzese2024continuous, hagemann2023multilevel}. Being used for generative modelling, these methods are limited to linear SDEs, i.e. SDEs with state-independent diffusion coefficients. Such a linear setting simplifies the construction of bridges as the transition probability is Gaussian and tractable. In a more general nonlinear case, where the diffusion coefficients are state-dependent, simulating diffusion bridges is nontrivial due to inaccessible closed form of the score functions.

\textbf{Diffusion Schrödinger bridge}: 
The diffusion bridge simulation problem studied in this paper is distinct from the diffusion Schrödinger bridge (DSB) problem, despite their similar names. A diffusion bridge describes a stochastic process conditioned to start at a fixed point and reach another fixed point at a given terminal time. In contrast, the DSB problem seeks a stochastic process that transports an initial probability distribution to a target distribution in a way that minimizes a relative entropy functional, making it a form of entropy-regularized optimal transport. While DSB methods have gained popularity in generative modeling \citep{de2021diffusion, shi2024diffusion, tang2024simplified, thornton2022riemannian}, it is crucial to distinguish between these two problems, as they involve fundamentally different formulations and objectives.

\subsection{Contributions}
We aim to incorporate score matching techniques into the diffusion bridge simulation methodology, especially in handling infinite-dimensional nonlinear bridges. Our contributions are outlined as follows:
\begin{itemize}
    \item We derive the time reversal of the infinite-dimensional diffusion bridge, together with a computable optimization objective under finite discretization.
    \item We design a time-dependent neural operator structure that can learn the infinite-dimensional object through finite samples.
    \item We demonstrate our method with various continuous function-data-valued stochastic processes and compare them against other related methods.
\end{itemize}

\section{PRELIMINARIES}
\subsection{Doob's \texorpdfstring{\(h\)}{h}-transform}
Given a \emph{finite-dimensional} Euclidean diffusion process \(\bX = \{\bX_t\}_{t\in[0, T]}\in \R^d\) governed by the SDE:
\begin{equation}    \label{eq: f_sde}
    \d \bX_t = \mathbf{f}(t, \bX_t)\d t + \mathbf{g}(t, \bX_t)\d \mathbf{W}_t
\end{equation}
with \(\mathbf{f}:[0, T] \times \R^d \to \R^d, \mathbf{g}:[0, T] \times \R^d \to \R^{d'}\) as finite-dimensional drift and diffusion terms and \(\mathbf{W}\) as a Wiener process in \(\R^{d'}\). When the event \(\bX_T=\mathbf{v}\) is observed at \(t=T\), the dynamics of \(\bX\) changes according to Doob's \(h\)-transform \citep{rogers2000diffusions} by reweighting the transition probabilities. \cite{baker2024conditioning} showed that such a transformation can be generalized to \emph{infinite dimensions} as well.

Let \((\H,\langle\cdot,\cdot\rangle_{\H})\) denote a separable Hilbert space with a chosen countable orthonormal basis \(\{e_i\}_{i\in\mathbb Z}\).
Let \((\Omega, \mathcal{F}, \mathbb{P})\) be a probability space with natural filtration \(\{\mathcal{F}_t\}\). 
An \(\H\)-valued diffusion process \(X=\{X_t\}_{t\in[0, T]}\) is defined as: 
\begin{equation} \label{eq: inf_sde}
    \d X_t = f(t, X_t)\d t + g(t, X_t)\d W^{\P}_t,\quad X_0=u\in\H
\end{equation}
where \(W^{\P}\) is a \(\P\)-Wiener process with a covariance operator \(Q\) in a Hilbert space \(\U\) (where \(\U\) can equal \(\H\)), \(f:[0, T]\times \H \to \H\), \(g:[0, T]\times \H\to\operatorname{HS}(Q^{1/2}(\U), \H)\), where \(\operatorname{HS}(Q^{1/2}(\U), \H)\) denotes the Hilbert-Schmidt operator \(Q^{1/2}(\U)\to\H\). The following theorem \citep[Theorem 5.1]{baker2024conditioning} states a change of dynamics of \(X\):
\begin{theorem} \label{thm: inf_doob}
    Let \(h:[0, T]\times\H\to\R_{>0}\) be a continuous Fréchet differentiable function. Let \(Z(t)\coloneq h(t, X_t)\) be a strictly positive mean-one martingale. Define a new measure \(\mathbb{P}^{\star}\) by:
    \begin{equation}
        \d \mathbb{P}^{\star} \coloneq Z(T)\d \mathbb{P}.
    \end{equation}
    Then under the new defined measure \(\mathbb{P}^{\star}\), \(X_t\) solves
    \begin{multline} \label{eq: inf_doob_sde}
        \d X_t = f^{\star}(t, X_t)\d t + g(t, X_t)\d W^{\Ps}_t ,\quad X_0=u, \\
        f^{\star}(s, x) = f(s, x) + a(s, x)\nabla_x\log h(s, x),
    \end{multline} 
    where \(W^{\Ps}\) is a \(\mathbb{P}^{\star}\)-Wiener process. Let \(g^*:[0,T]\times\H\to\operatorname{HS}(\H, Q^{1/2}(\U))\) be the adjoint operator of \(g\), \(a=gg^*:[0, T]\times \H\to \operatorname{HS}(\H, \H)\), and \(\nabla_x\log h :[0, T]\times\H\to\H\) be the score function. We further denote \(X^{\star}\) as \(X\) under the measure \(\mathbb{P}^{\star}\).
\end{theorem}
\cref{thm: inf_doob} reveals a change of SDE by a change of measures. Moreover, \cite{baker2024conditioning} demonstrated that when such a change is induced by conditioning on \(X_T\in\Gamma\subseteq \H\), \(h(s, x) = \mathbb{P}(X_T\in\Gamma\mid X_s=x)\). However, \cref{eq: inf_doob_sde} is not directly available to sample because 1) \(\nabla_x\log h(s, x)\) is inaccessible; 2) it is infinite-dimensional. We will address these two issues in \cref{sec: methodology} by applying a score matching technique on the infinite-dimensional time-reversal process and a finite basis projection.

\subsection{Time-reversed diffusion processes} \label{sec: time_reversed_diffusion_processes}
\cite{haussmann1986time} showed that under certain conditions (see \cite[Assumption (A)]{haussmann1986time}), the finite-dimensional diffusion process \(\bX\) has a time-reversed process \(\bY\), which solves the following SDE:
\begin{multline} \label{eq: finite_t_reversed_sde}
    \d \bY_t = \bar{\mathbf{f}}(T-t, \bY_t)\d t + \mathbf{g}(T-t, \bY_t)\d \mathbf B_t,\\
    \bar{\mathbf{f}}(T-s, x) = - \mathbf{f}(s, x) + p^{-1}(x)\nabla_x\left(\mathbf{a}(s, x)p(x)\right),
\end{multline}
where $\mathbf B_t$ is another Wiener process, \(\mathbf{a}(s, x) = \mathbf{g}\mathbf{g}^{\intercal}(s, x)\), and \(p(x)\) is the density of \(\bX\). 

When lifting to infinite dimensions, \cite{millet1989time} showed that the projections of \(X\) also have a system of similar time-reversed SDEs. To show that, let \(\{k_j\}_{j=1}^\infty\) denote a countable orthonormal basis of \(\U\). Write \cref{eq: inf_sde} as an infinite system of \(\mathbb{R}\)-valued SDEs with respect to the basis \(\{e_i\}_{i=1}^\infty\) of \(\H\):
\begin{equation} \label{eq: dis_inf_sde}
    \d [X_t]_{i} = [f(t, X_t)]_i\d t + \sum_{j}^{\infty} [g(t, X_t)]_{ij}\d [W^{\P}_t]_j,
\end{equation}
 where \([X_t]_i, [W^{\P}_t]_j\) are the \(i\)-th, \(j\)-th components of \(X_t, W^{\P}_t\) under the bases \(e_i, k_j\), defined by the inner products \(\langle\cdot,\cdot\rangle_{\H}, \langle\cdot,\cdot\rangle_{\U}\), \([f(s, x)]_i:=\langle f(s, x), e_i\rangle_{\H}\), \([g(s, x)]_{ij}:= \langle g(s, x)(k_j), e_i\rangle_{\H}\). 

\cite{millet1989time} made an analogy to the finite-dimensional time reversal diffusion. Suppose that given \(i\in\mathbb{Z}\), there exists a \emph{finite} \(I(i)\subset\mathbb{Z}\), such that for all \(s\), \([a(s, x)]_{ij}\coloneq\sum_\ell^{\infty}[g(s,x)]_{i\ell}[g(s,x)]_{j\ell}=0\) if \(j\notin I(i)\). Intuitively, \(I(i)\) contains the indices of component $[X]_i$ that are conditionally dependent. Denote \(\bx{i}:= \{[X]_j;j\in I(i)\}\in\R^{|I(i)|}\) and \(\hat{\mathbf{x}}^{(i)}:= \{[X]_j;j\notin I(i)\}\). For \(t>0\), and \(\bz{i}:=\{z_j;\,z_j\in\R,j\notin I(i)\}\), the conditional law of \(\bx{i}\) given \(\hat{\mathbf{x}}^{(i)}=\bz{i}\) is assumed to have a density \(p(\bx{i}\mid\bz{i})\) (see \cite{millet1989time} Theorem 5.3 for the restrictions on the coefficients that guarantee the existence of the density). Then \cite{millet1989time} Theorem 3.1 gives the time reversal of \cref{eq: dis_inf_sde} as:
\begin{multline}
    \d [Y_t]_i = [\bar{f}(T-t, Y_t)]_i\d t + \sum_{j}^{\infty} [g(T-t, Y_t)]_{ij}\d [W^{\hat \P}_t]_j, \\
    [\bar{f}(T-s, x)]_i = -[f(s, x)]_i \\
    + p^{-1}(\bx{i}|\bz{i})\sum_{j\in I(i)}\nabla_j\left([a(s, x)]_{ij}p(\bx{i}|\bz{i})\right).
\end{multline}
$\hat \P := \P\circ \varphi^{-1}_T$ is the push-forward of $\P$ by the time-reversal map $\varphi_t:\Omega\to\Omega$ defined by $\varphi(\omega)(t):=\omega(T-t), \omega \in \Omega $. This result is especially useful, as it shows that we can find a well-defined time reversal for an infinite-dimensional diffusion process in terms of the conditional density in the finite subspace of \(\H\), which shall serve as the foundation of our infinite-dimensional time-reversed diffusion bridge.

\subsection{Denoising score matching}
Score matching \citep{hyvarinen2005estimation} is a technique that is used to estimate the score function \(\nabla_x\log p(x_t)\) by a parameterized estimator \(s_{\theta}(t, x)\). An important variant of score matching is denoising score matching (DSM) \citep{vincent2011connection}, where the optimization problem is formulated as:
\begin{equation} 
    L_{\text{DSM}}(\theta) = \E_{x}\left[\|s_{\theta}(t, x) 
    - \nabla_x\log p(x\mid x_0)\|^2\right],
\end{equation}
where \(p(x|x_0)\) as the transition density from \((0, x_0)\) to \((t, x)\). The DSM loss requires \(X\) to be linear with Gaussian transition density \(p(x|x_0)\). For more general diffusion processes, inspired by DSM, \cite{heng2021simulating} introduced a variational formulation to learn the intractable \(\nabla_x\log p(x\mid x_0)\), by defining the optimization problem:
\begin{gather}
    L_{\text{t-reversed}}(\theta) = \frac{1}{2}\sum^{M}_{m=1}\int^{t_m}_{t_{m-1}}\E_{x}\left[\|s_{\theta}(t, x) \right. \nonumber\\
    \left.- \nabla_x\log p(x \mid x_{t_{m-1}})\|^2_{\mathbf{a}(t, x)}\right]\d t.
\end{gather}
They propose applying this optimization twice on forward and backward processes sequentially to recover the original forward-in-time conditional process.

\section{METHODOLOGY} \label{sec: methodology}

\subsection{Infinite-dimensional time-reversed bridge}
To extend the framework proposed by \cite{heng2021simulating} into infinite-dimensional settings, we need to establish the time reversal of the infinite-dimensional diffusion bridge, which is given by the following theorem:

\begin{theorem}     \label{thm: t_reversal}
Let \(f, g, a\), and \(W^{\P}_t\) be as defined before. The conditional process \(\Xs\) has a time reversal \(\Ys = \{\Ys_t\}_{t\in[0,T]} = \{\Xs_{T-t}\}_{t\in[0,T]}\), with the chosen bases \(e_i\) and \(k_j\) of \(\H\) and \(\U\) respectively, which follows the SDEs:
\begin{equation} \label{eq: inf_r_b_sde}
    \d [\Ys_t]_i = [\overline{f^{\star}}(t, \Ys_t)]_i\d t + \sum^{\infty}_{j}[g(T-t, \Ys_t)]_{ij} \d [W^{\P}_t]_j
\end{equation}
where, 
\begin{multline} \label{eq: dis_r_b_drift}
    [\overline{f^{\star}}(T-s, x)]_i = -[f(s, x)]_i + \sum_{j\in I(i)}\nabla_{j}[a(s,x)]_{ij} \\
    + \sum_{j\in I(i)}[a(s, x)]_{ij}\nabla_{j} \log p(\bx{i}\,|\, (\bz{i}, x_0))
\end{multline}
\end{theorem}

The proof of \cref{thm: t_reversal} can be found in \cref{sec: t_reversal_thm_proof}. A crucial insight is that $Y^{\star}$ and $X$ share the same path measure $\P$, and sampling from $\P$ is trivial by sampling $X$ instead of $Y^{\star}$. Note that \cref{eq: dis_r_b_drift} involves \(p(\bx{i}\,|\, (\bz{i}, x_0))\), which is the conditional density of \(\bx{i}\) given \(\hat{\mathbf{x}}^{(i)}=\bz{i}\) and \(X_0=x_0\in\H\). Compared with the \(h\)-function in \cref{eq: inf_doob_sde}, we are able to design an objective function to learn this finite-dimensional density. Inspired by \cite{heng2021simulating}, we use a variational approach to approximate the unconditioned time reversal \(Y\) with a parameterized operator \(\mathcal{G}^{(\theta)}(s, x):[0, T]\times\H\to\H\), such that when the object function is optimized, its projection under \(e_i\), \([\mathcal{G}^{(\theta)}(s, x)]_i\), approximates \(\sum_{j\in I(i)}[a(T-s, x)]_{ij}\nabla_{j} \log p(\bx{i}\,|\, (\bz{i}, x_0))\), and therefore is available in \cref{eq: inf_r_b_sde}.

\subsection{Learning the time reversal} \label{sec:loss_function}
To approximate \(Y\), we define a diffusion process \(Y^{(\theta)}=\{Y_t^{(\theta)}\}_{t\in[0, T]}\in\H\) that solves:
\begin{equation} \label{eq: y_approximate}
    \d \Ytheta_t = f^{(\theta)}(T-t, \Ytheta_t)\d t + g^{(\theta)}(T-t, \Ytheta_t)\d W^{\Pt}_t,
\end{equation}
where \(f^{(\theta)}:[0, T]\times \H\to \H\) is a mapping parameterized by \(\theta\in\Theta\), and \(W^{\Pt}\) is a \(\Pt\)-Brownian motion. The set of measures \(\{\Pt;\;\theta\in\Theta\}\) provides a variational class for approximating \(\P\). The absolute continuity between \(\Pt\) and \(\P\) is guaranteed by choosing \(g^{(\theta)}(T-s, x) = g(T-s, x)\). Furthermore, we formulate \(f^{(\theta)}\) such that its projection under \(e_i\) satisfies:
\begin{multline} \label{eq: f_theta_component}
    [f^{(\theta)}(T-s, x)]_i = -[f(s, x)]_i \\
    + \sum_{j\in I(i)}\nabla_{j}[a(s, x)]_{ij} + [\mathcal{G}^{(\theta)}(s, x)]_i, 
\end{multline}
where \(\mathcal{G}^{(\theta)}(t, x):[0, T]\times \H \to \H\) is a bounded parameterized operator. The advantage of such a formulation is that the KL divergence \(\kld(\P||\Pt)\) can be expressed by the difference between the drift coefficients explicitly, obtained by the following lemma:
\begin{lemma} \label{lem: kld}
    Let \(\{e_i\}\) be the basis for which \(a(s, x)\) is diagonalizable for any \(s\in[0, T], x\in\H\). Let \(f^{(\theta)}:[0,T]\times\H\to\H\) be a Lipschitz function of linear growth, parameterized by \(\theta\), such that its decomposition under \(e_i\) follows \cref{eq: f_theta_component}. Define the difference of drifts as:
    \begin{equation}
        [\Psi(s, x)]_i \coloneq [\{\bar{f} - f^{(\theta)}\}(s, x)]_i.
    \end{equation}
    Then the KL divergence can be expressed as:
    \begin{equation} \label{eq: kld}
        \mathrm{D}_{\operatorname{KL}}(\P||\Pt) =  \frac{1}{2}\E_{\P}\left[\int^{T}_{0}\sum^{\infty}_{i=1} \lambda_i[\Psi(s, X_s)]_i^2\,\d s\right],
    \end{equation}
    where \(\lambda_i=\|g^*(e_i)\|^2_{Q^{1/2}(\U)}\) is the \(i\)-th eigenvalue of \(a\), and \(\Psi(s, x)=\sum_{j\in I(i)}[a(s, x)]_{ij}\nabla_{j}\log p(\bx{i}\,|\,\bz{i}) - [\mathcal{G}^{(\theta)}(s, x)]_i\).
\end{lemma}
It turns out that matching the two measures via minimizing the KL divergence is equivalent to matching the two drifts by their components. The proof can be found in \cref{sec: kld_lemma_proof}. However, \cref{eq: kld} still cannot be used as the optimization object due to 1) the intractable \(p(\bx{i}|\bz{i})\); 2) the infinite sum. Fortunately, since \(X\) is Markovian, it is possible to work with a smaller transition step \(t_{n-1}\to t\) such that \(t-t_{n-1}\ll T\). The following theorem gives an objective that shares the same gradient with respect to \(\theta\) as \(\kld(\P||\Pt)\). The proof of theorem is shown in \cref{sec: loss_thm_proof}.

\begin{theorem}\label{thm: loss}
    For any time partition \(\{t_n\}_{n=1}^{N}\) of the interval \([0, T]\) and arbitrarily chosen orthogonal basis \(e_i\) of \(\H\), define
    \begin{equation}
        [\Psi'(s, x; x_{t_{n-1}})]_i = [\mathcal{G}^{(\theta)}(s, x)]_i - [b_s(x; x_{t_{n-1}})]_i,
    \end{equation}
    with \(b_s(\cdot\;; x_{t_{n-1}}):\H\to\H\) for \(x_{t_{n-1}}\in\H\), such that its component under \(e_i\) satisfies \([b_s(x; x_{t_{n-1}})]_i = \sum_{j\in I(i)}[a(s, x)]_{ij}\nabla_j\log p(\bx{i}\mid(\bz{i},  x_{t_{n-1}}))\). Then the objective function
    \begin{equation} \label{eq: loss}
         L(\theta) = \frac{1}{2}\sum^{N}_{n=1}\int^{t_n}_{t_{n-1}}\mathbb{E}_{\P}\left[ \sum_{i=1}^{\infty}\lambda_i[\Psi'(s, X_s; x_{t_{n-1}})]_i^2\right]\d s
    \end{equation}
    is equivalent to \(\mathrm{D}_{\operatorname{KL}}(\P||\Pt)\) up to a \(\theta\)-independent constant \(C\), where \(p(\bx{i}\mid(\bz{i},  x_{t_{n-1}}))\) denotes the conditional density of \(\bx{i}\) at time \(s\) given \(\hat{\mathbf{x}}=\bz{i}, X_{t_{n-1}}=x_{t_{n-1}}\in\H\), with \(\bx{i}, \hat{\mathbf{x}}^{(i)}, \bz{i}\) as defined before. \(X_s\) is sampled from \cref{eq: inf_sde}.
\end{theorem}

\subsection{Sampling the infinite-dimensional object}
To practically address the infinite sum, we truncate the bases \(e_i\) and \(k_j\) up to a finite \(M\), and set \(\bz{i}=\{[X]_j;j>M\}=0\). In such a setting, \(b_s(x; x_{t_{n-1}})\) can be approximated by:
\begin{equation}
    [\hat{b}_s(x; x_{t_{n-1}})]_i \approx \sum_{j=1}^{M}[a(s, \tilde{x})]_{ij}\nabla_{j}\log p(\tilde{x}\mid \tilde{x}_{t_{n-1}}),
\end{equation}
where \(\tilde{x} = \{[x]_1,\cdots,[x]_M\}\in\R^M\). Although \(p(\tilde{x}\mid \tilde{x}_{t_{n-1}})\) is generally intractable, we can approximate it by a Gaussian transition when using a discrete numerical solving scheme like Euler-Maruyama, as long as the time step \(\Delta t = s - t_{n-1}, s\in[t_{n-1}, t_n]\) is small enough. Specifically, we can simulate a finite system of \cref{eq: dis_inf_sde} containing \(M\) \(\R\)-valued SDEs:
\begin{equation}
    \d [X_t]_{i} = [f(t, X_t)]_i\d t + \sum_{j}^{M} [g(t, X_t)]_{ij}\d [W^{\P}_t]_j,
\end{equation}
and then,
\begin{multline}
    [\hat{b}_s(x; x_{t_{n-1}})]_i = \sum_{j=1}^{M}[a(s, \tilde{x})]_{ij}\nabla_j\log p(\tilde{x} \mid \tilde{x}_{t_{n-1}}) \\
    \approx -(\Delta t)^{-1}\cdot([\tilde{x}]_i-[\tilde{x}_{t_{n-1}}]_i - \Delta t\cdot [f(t_{n-1}, x)]_i) \\
    =-(\Delta t)^{-1}\cdot\sum^M_{j=1}[g(t_{n-1}, x)]_{ij}\Delta W,
\end{multline}
for $\Delta W\sim \mathcal N(0, \Delta t)$. Finally, we set both bases to be Fourier bases and compute \([X_t]_i\) by applying the Fast Fourier Transform (FFT) on the evaluation of \(X\) on a finite spatial-temporal grid \(\{\xi_i\}_{i=1}^M\times \{t_n\}_{n=1}^N\), that is, at each time step $t_n$, we implicitly choose the basis as the Fourier basis, then the decomposition $[\mathcal G^{(\theta)}]_i=\mathcal{F}[\mathcal G^{(\theta)}(X_{t_n}[\xi_i])]$. We then incorporate the transformation into $\mathcal G^{(\theta)}$ and denote the composition $\mathcal F\circ \mathcal G^{(\theta)}$ still as $\mathcal G^{(\theta)}$. In the following section, we shall discuss the design of \(\mathcal{G}^{(\theta)}\) such that it incorporates the FFT within its structure and, therefore, only accepts the evaluation of functions as input. We define our optimization objective as:
\begin{multline}    \label{eq: final_loss}
    L(\theta) = \frac{1}{2}\sum^{N}_{n=1}\int^{t_n}_{t_{n-1}}\mathbb{E}_{\P}\left[ \sum_{i=1}^{M}\lambda_i\Big( \mathcal{G}^{(\theta)}(s, X_s[\xi_i]) \right.\\
    \left.+ (\Delta t)^{-1}\cdot \sum^M_{j=1}[g(t_{n-1}, x)]_{ij}\Delta W\Big)\right]\,\d t.
\end{multline}
Under the finite truncation, \(a\) can be represented as a \(M\times M\) matrix \(a_{M\times M}\), and since \(a\) must be diagonalizable under \(e_i\), \(\lambda_i\) can be simply chosen as the diagonal entries of \(a_{M\times M}\).

\subsection{Time-dependent Fourier neural operators}

\subsubsection{Fourier neural operator}
\begin{figure*}[!ht] 
  \centering
  \includegraphics[width=0.8\textwidth]{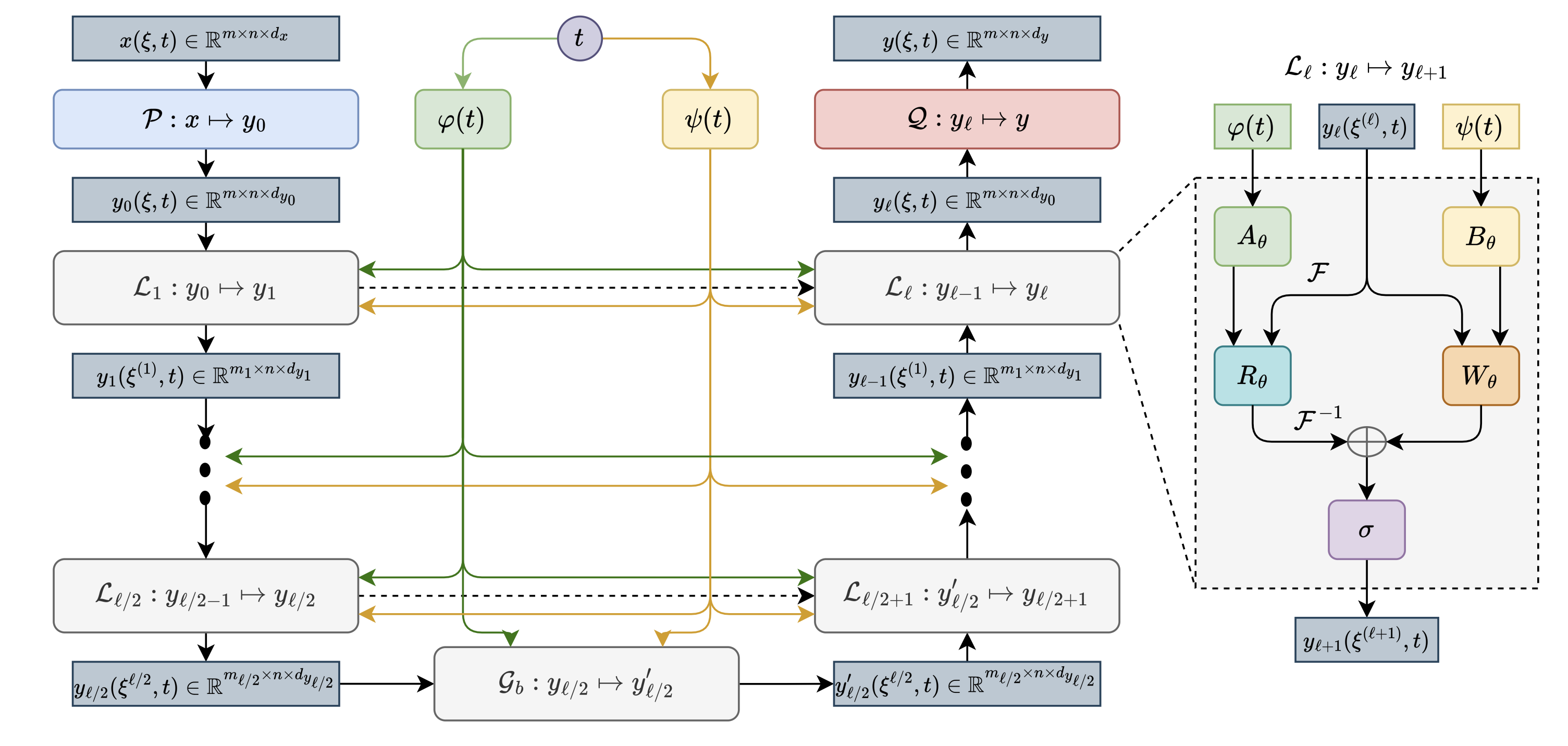} 
  \caption{Continuous-time U-shaped FNO architecture, acting as a parameterized operator \(\mathcal{G}^{(\theta)}:(x, t)\mapsto y\) for \(x,y\in \H\). Both \(x, y\) are evaluated on a discrete spatial-temporal grid \((\xi, t)\) with \(\xi\in\R^{m^d\times d}\) and \(t\in\R^n\). The figure only shows the case when \(d=1\), but the application to \(d>1\) can be achieved through similar architecture, e.g. \(x(\xi, t)\in\R^{m\times m\times n\times d_x}\) for \(d=2\) and so on.}
  \label{fig: ctuno}
\end{figure*}
As described in the previous section, \(\mathcal{G}^{\theta}(t, x):[0, T]\times \H\to \H\) is a time-dependent operator. \cite{franzese2024continuous} proposed two different approaches to model it: implicit neural representations \citep{saharia2022photorealistic} and Transformers \citep{vaswani2017attention}. Another solution suggested by \cite{baker2024conditioning} is to choose a certain truncated basis and use a multi-layer perceptron (MLP) to learn the coefficients under the projection. We found that compared with these tools, neural operators \citep{li2020fourier, li2020neural, lu2021learning, kovachki2023neural} offer a more natural representation of \(\mathcal{G}^{\theta}(t, x)\), as it directly parameterizes a nonlinear operator in a learnable fashion. 
Specifically,
\begin{equation}
    \mathcal{G}^{(\theta)}:= \mathcal{Q}\circ \mathcal{L}_{L}\circ\dots\circ\mathcal{L}_{1}\circ\mathcal{P},\quad x\mapsto y
\end{equation}
where \(x\in\H(\Omega;\R^{d_x}), y\in\H(\Omega;\R^{d_y})\), \(\mathcal{P}:\R^{d_x}\to\R^{d_1}, \mathcal{Q}:\R^{d_L}\to\R^{d_y}\) are the local lifting and projection mappings respectively. 
The time-independent Fourier layer \citep{li2020fourier} \(\mathcal{L}_{\ell}:\H(\Omega;\R^{d_{x}})\to \H(\Omega;\R^{d_{\ell}})\) is defined as:
\begin{equation}
    \mathcal{L}_{\ell}(x)(\xi) := \sigma\left(W_{\ell}(x)(\xi) + \mathcal{F}^{-1}\{R_{\ell}\cdot(\mathcal{F}\{x\})\}(\xi)\right),
\end{equation}
where \(\xi=(\xi_1, \dots, \xi_m)\in\R^m\) is the finite grid discretization of the domain \(\Omega\),
\(W_{\ell}\) is a linear transform \(\R^{d_x}\to\R^{d_{\ell}}\), 
and \(\mathcal{F}\{\cdot\}, \mathcal{F}^{-1}\{\cdot\}\) are the FFT and inverse FFT (iFFT) respectively, \(R_{\ell} = \mathcal{F}\{\kappa_{\ell}\}:\R^m\to\mathbb{C}^{d_{\ell}\times d_{\ell}}\) for kernel \(\kappa_{\ell}:\R^m\to\R^{d_{\ell}\times d_{\ell}}\), and \(\sigma:\R\to\R\) is the nonlinear activation function applied component-wise. 

\subsubsection{Continuous-time modulation}
In \cite{li2020fourier}, the designed \(\mathcal{G}^{(\theta)}\) is time-independent, even though the domain \(\Omega\) can be the product of \([0, T]\subset\R\) with some coordinate domain. While in our case, \(\mathcal{G}^{(\theta)}\) depends on \(t\) explicitly. A natural idea is to introduce additional time embeddings into the weight matrix \(W_{\ell}(x)\) at each layer, i.e., \(W'_{\ell}(t, s)=\alpha(t)W_{\ell}(x)+\beta(t)\), where \(\alpha, \beta\) are learnable MLPs to embed \(t\). However, such a setting ignores the contribution from the frequency component \(\mathcal{F}(x)\). To address this, \cite{park2023learning} proposed a continuous-time-modulated Fourier layer formulation to integrate such a time dependency into both physical and frequency domains:
\begin{multline}
    \mathcal{L}_{\ell}(x)(t, \xi) = \sigma\left(W_{\ell}\psi_{\ell}(t)x(\xi) \right.\\
    \left.+ \mathcal{F}^{-1}\{\varphi_{\ell}(t)R_{\ell}\cdot(\mathcal{F}\{x\})(\xi)\}\right)
\end{multline}
where \(\psi_{\ell}(t)\in\R^{d_{\ell}\times d_{\ell}}\), \(\varphi(t):\R^m\to\mathbb{C}\) are time modulations. Essentially, the time information is not only modulated into the physical domain through \(\psi(t)\), but also into the Fourier domain through \(\varphi(t)\). 

In practice, \(\psi(t)\) and \(\varphi(t)\) are implemented with the same structure, as suggested in \cite{park2023learning}, by using the sinusoidal embeddings \citep{vaswani2017attention}. We find introducing the time modulation in both the physical and frequency domains shows better performance than only acting on the physical domain using transformed time embeddings to shift and scale the output from the Fourier block. 

\subsubsection{U-shaped FNO}
The operator \(\mathcal{G}^{(\theta)}(t, \cdot)\) maps the element \(x\in\H\) into the same space \(\H\), preserving the dimensionality of the input. This structure-preserving property motivates the adoption of a U-shaped architecture for the Fourier Neural Operator (FNO), as proposed by \cite{ashiqur2022u}. The U-shape design is particularly advantageous for: 1) extracting and refining multi-level Fourier features across successive scales, enabling the operator to capture both local and global dependencies efficiently; 2) reducing redundant parameterization by reusing feature maps at corresponding resolutions, thereby lowering memory costs while maintaining expressive power.

We combine the continuous-time modulation and U-shape skeleton, and propose the \emph{continuous-time U-shaped FNO}, as shown in \cref{fig: ctuno}. This operator is used to map the pair \((x\in \H, t\in[0, T])\) onto \(y\in\H\). We assume that the input and output functions share the same spatial and temporal domain, and the spatial domain is usually set as a \(d\)-dimensional manifold \(\mathcal{M}\). We choose the spatial domain as periodic subsets of \(\R\) or \(\R^2\). In the experiment section, we shall detail the choice of such bounded domains for specific tasks. 

To evaluate the functions, the spatial domain is discretized into \(m\times\dots\times m=m^d\) grid points in \(\R^d\), and the temporal domain \([0, T]\) is also discretized into \(n\) intervals. Then the evaluation of \(x\), \(x(\xi, t)\in\R^{m\times\dots\times m\times n\times d_x}\), where \(d_x\) is the dimension of the codomain of \(x\). For example, for a function \(x\in\H([0, T]\times\Omega, \R), \Omega\subset\R\), its evalutation can be represented as the tensor of shape \([m, n, 1]\).

The lifting operator \(\mathcal{P}\) acts on the codomain which lifts its dimension to \(d_{y_0}\). The lifted representation together with the \(\psi(t), \varphi(t)\) embeddings are fed into one of the continuous-time Fourier layers \(\mathcal{L}_{\ell}\), where inside the layer, \(\varphi(t)\) and \(\psi(t)\) are transformed by independent learnable linear layers \(A_{\theta}\) and \(B_{\theta}\) individually to fit the size of the Fourier and physical domain features. 
After the modulation, features from both domains are added and activated by the nonlinear activation function \(\sigma\). 

The FFT's inherent flexibility enables dynamic adjustment of spatial resolution through downsampling or upsampling, mirroring the multi-scale feature extraction in standard UNet architectures. This contrasts with conventional FNO implementations that maintain fixed intermediate resolution, as our U-shaped design facilitates simultaneous capture of low- and high-frequency features through multi-resolution processing. We preserve UNet-style skip connections by concatenating physical-domain features from each downsampling stage with corresponding upsampling outputs along the channel dimension, thereby doubling the codomain dimension for subsequent operations. This modified architecture outperforms baseline FNO implementations through three key advantages: (1) superior performance with fewer parameters, (2) faster convergence rates, and (3) enhanced suitability for operator approximation in diffusion bridge problems.

\section{EXPERIMENTS}

\subsection{Functional Brownian bridges}
Although the main motivation of our proposed method is for nonlinear processes, we spend some time evaluating cylindrical Brownian motion in various settings. This is because in this case we also know the true bridge processes, and therefore can fully evaluate the method against the truth.
For a Hilbert space \(\mathcal{U}\) with basis \(k_j\), the cylindrical Brownian motion can be defined as \(\sum_{i=1}^\infty w_i(t)e_i(t)\), where \(w_i\) are independent 1-dimensional Brownian motions. This sum does not converge in \(\mathcal{U}\), but can be shown to converge in another Hilbert space \citep{da2014stochastic}.

\begin{figure}[!ht]
    \centering
    \begin{subfigure}[b]{0.48\columnwidth}
        \includegraphics[width=\textwidth]{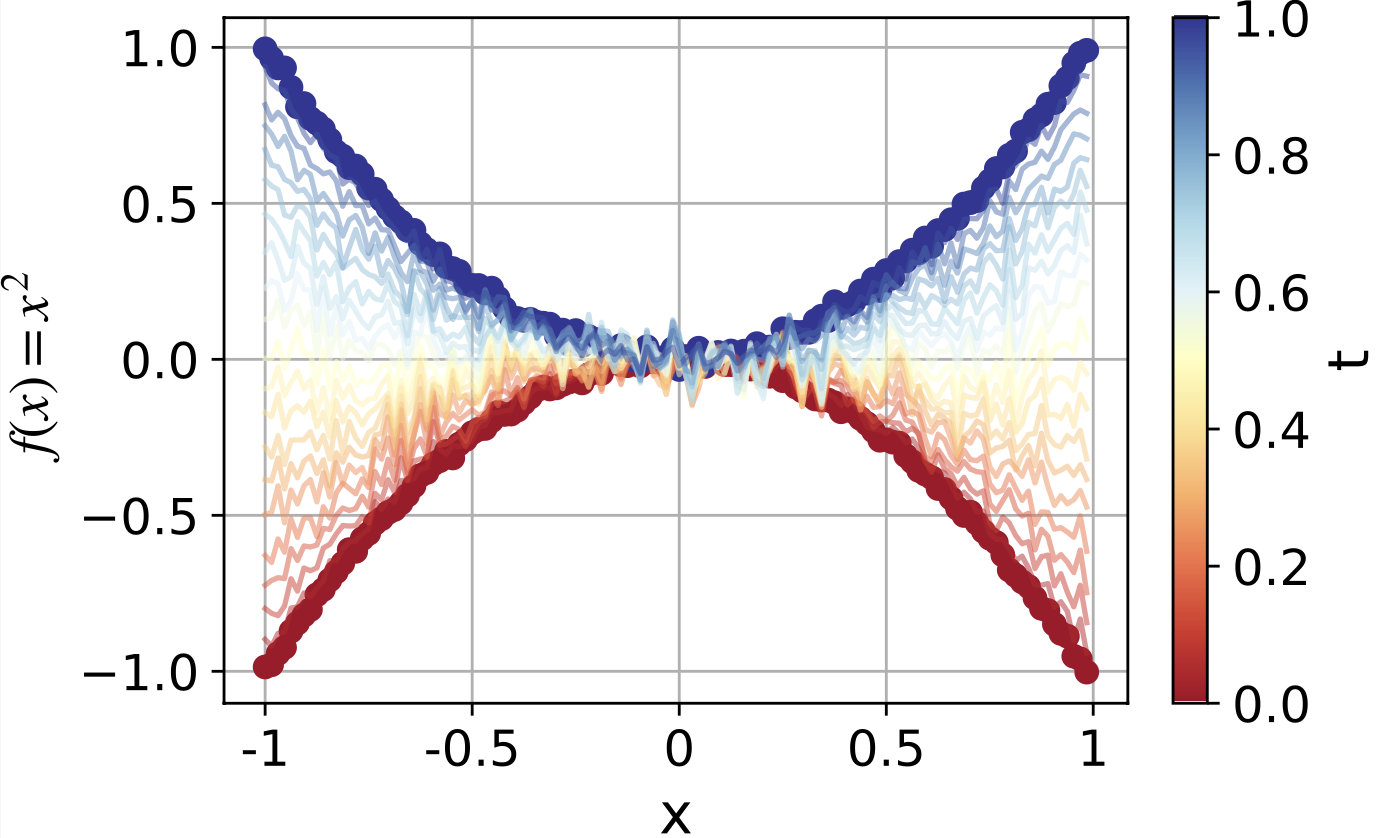}
        \caption{Learned bridge}
        \label{fig: 1(a)}
    \end{subfigure}
    \begin{subfigure}[b]{0.48\columnwidth}
        \includegraphics[width=\textwidth]{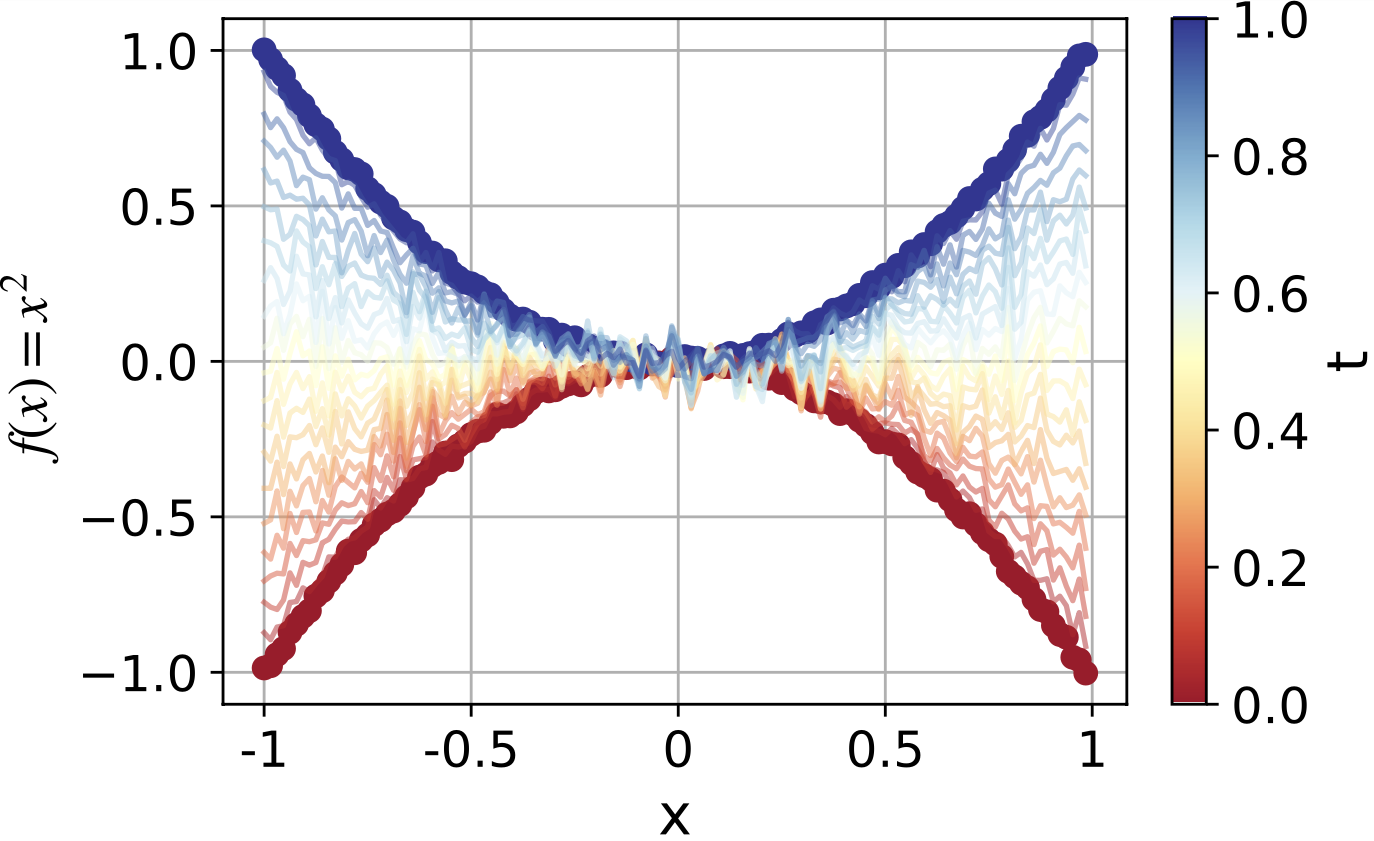}
        \caption{Ground truth}
        \label{fig: 1(b)}
    \end{subfigure}

    \caption{Qualitative results for Brownian bridges between two quadratic functions; (a) One sample from the learned reversed Brownian bridge, evaluated at 128 evenly distributed points; (b) One sample from the true reversed Brownian bridge, simulated with the same random seed as (a) for comparison.}
\end{figure}

\textbf{Quadratic functions:} We first study the cylindrical Brownian motion of quadratic function evolutions, whose score has a closed form. 
For the start/target functions, we choose \(f(x) = ax^2 + \varepsilon\), where \(a=\{1,-1\}\) and \(\varepsilon\sim \mathcal{N}(0, 10^{-4})\) is used to equip the set with non-zero measure, as used in \cite{phillips2022spectral}.
We choose the bounded grid of \([-1, 1]\) to evaluate the function, and the grid is uniformly discretized into a finite number of points. 
Such a discretization is equivalent to choosing a finite basis and projecting it into finite dimensions.
However, the projected process should be consistent under arbitrary discretizations. 
We show it by training the neural operator under a low-resolution scheme (8 points distributed within \([-1, 1]\)) and evaluate it in a high-resolution case (128 points). 
Note that the learned bridge (\cref{fig: 1(a)}) shows high consistency with the true Brownian bridge (\cref{fig: 1(b)}), even on a much finer grid.

\begin{figure}[!ht]
    \centering
    \begin{subfigure}[b]{0.48\columnwidth}
        \includegraphics[width=\textwidth]{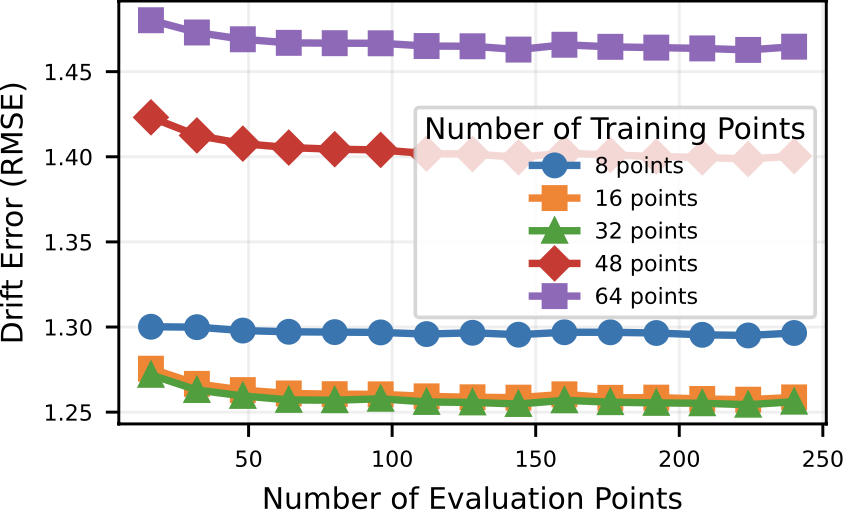}
        \caption{Drift error}
        \label{fig: 2(a)}
    \end{subfigure}
    \begin{subfigure}[b]{0.48\columnwidth}
        \includegraphics[width=\textwidth]{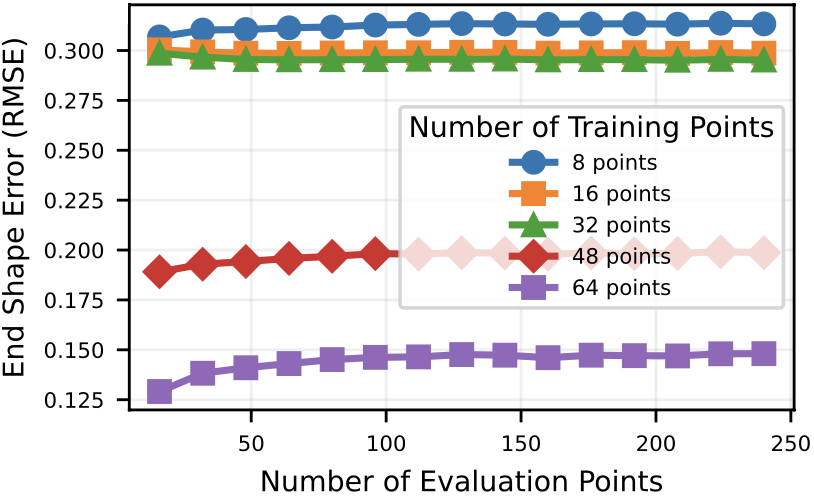}
        \caption{End shape error}
        \label{fig: 2(b)}
    \end{subfigure}

    \caption{Quantive evaluation of the model on different levels of discretization; (a) RMSE between the model's output and the ground true drift term evaluated on the whole trajectory; (b) RMSE between the end of estimated trajectories and the true target shape; All the statistics are done for 64 independent samplings.}
\end{figure}

\begin{figure}[!ht]
    \centering
    \includegraphics[width=1.0\columnwidth]{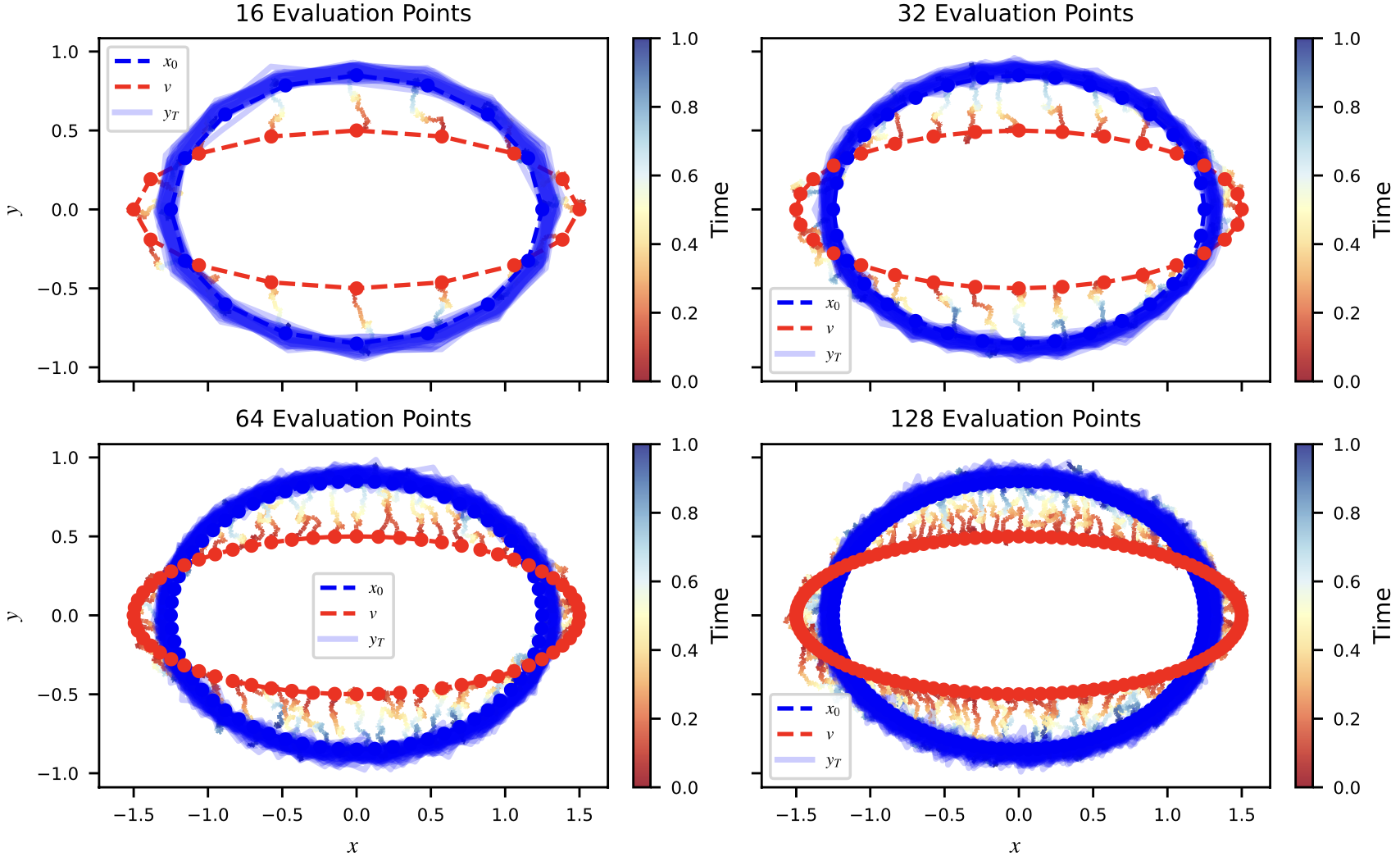}
    \caption{Visualization of Brownian bridges between ellipse shapes evaluated under different levels of discretizations, the training is done on 16 points. The blue shade represents 64 independent samples of end shape, and only one colored sample of the trajectories is shown with different colors indicating time steps.}
    \label{fig: 3}
\end{figure}

\textbf{Ellipses}: We further test our method by simulating the Brownian bridge between 2D ellipses, which acts as the simplest 2D stochastic shape matching task whilst still holding a known form to which we can compare.
An ellipse can be treated as a function from \([0, 1]\subset\R\) to \(\mathcal{S}^1\subset\R^2\), and can be characterized by finite points along the outlines. 
We are trying to bridge two ellipses with different axes.
As we expect our method to show consistency under different levels of discretization, we train the same model with different numbers of training sample points and evaluate it on more points. Then we compare the model output at all discrete time steps against the true drift term of the Brownian bridge. 
\cref{fig: 2(a)} and \cref{fig: 2(b)} show the results. 
The model demonstrates consistent generalization from finite training samples to infinite-dimensional processes, with errors converging under finer discretization. Fewer training points yield lower drift error but higher end shape discrepancies, as reduced dimensionality facilitates drift estimation while neglecting higher Fourier modes. This truncation implicitly zeros unlearned basis elements, producing over-smoothed trajectories that amplify terminal errors despite accurate drift recovery. \cref{fig: 3} visualizes these truncation effects in Brownian bridges between ellipses.

The neural operator demonstrates superior accuracy and memory efficiency for drift estimation compared to traditional coefficient-based approaches. To our knowledge, \cite{baker2024conditioning} represents the sole prior work addressing general infinite-dimensional diffusion bridges through truncated Fourier coefficient learning via MLPs. We establish a Brownian motion benchmark where closed-form solutions exist, unlike nonlinear scenarios where analytical solutions become intractable. \cref{tab: benchmark} compares the root mean squared error (RMSE) of drift estimates against ground truth across all time steps, contrasting our neural operator with the Fourier network in \cite{baker2024conditioning}. Notably, our operator achieves consistent error levels even with increased evaluation points (i.e., larger basis sets), whereas the Fourier network exhibits error escalation with higher spatial resolution. Furthermore, our architecture achieves comparable accuracy with substantially fewer parameters. Additional implementation details and results are provided in \cref{sec: app_experiment_results}.

\begin{table*}[ht]
    \centering
    \begin{tabular}{ccccccc}
       \toprule
         & \multicolumn{3}{c}{Neural operator}  & \multicolumn{3}{c}{Fourier coeffs net} \\
         & \multicolumn{3}{c}{(\#modes in 1st Fourier layer)} & \multicolumn{3}{c}{(\#truncated modes)} \\
       \#eval pts  & 8 & 16 & 32 & 8 & 16 & 32 \\
       \midrule
       32  & 1.5966 & 1.5894 & 1.5389 & 3.1133 &  5.7329 & 9.7032\\
       64  & 1.5959 & 1.5815 & 1.5271 & 3.1092 &  5.7279 & 9.7080\\
       128 & 1.5937 & 1.5779 & 1.5198 & 3.1085 &  5.7313 & 9.7389\\
       256 & 1.5984 & 1.5768 & 1.5142 & 3.1091 &  5.7317 & 9.7273\\
       \#params & 131,314 & 185,586 & 294,130 & 220,608 & 875,392 & 3,487,488 \\
       \bottomrule
    \end{tabular}
    \caption{Comparison between the neural operator and the MLP for Fourier coefficients, tested on Brownian bridges between ellipses.}
    \label{tab: benchmark}
\end{table*}

\begin{figure*}
    \centering
    \includegraphics[width=0.8\textwidth]{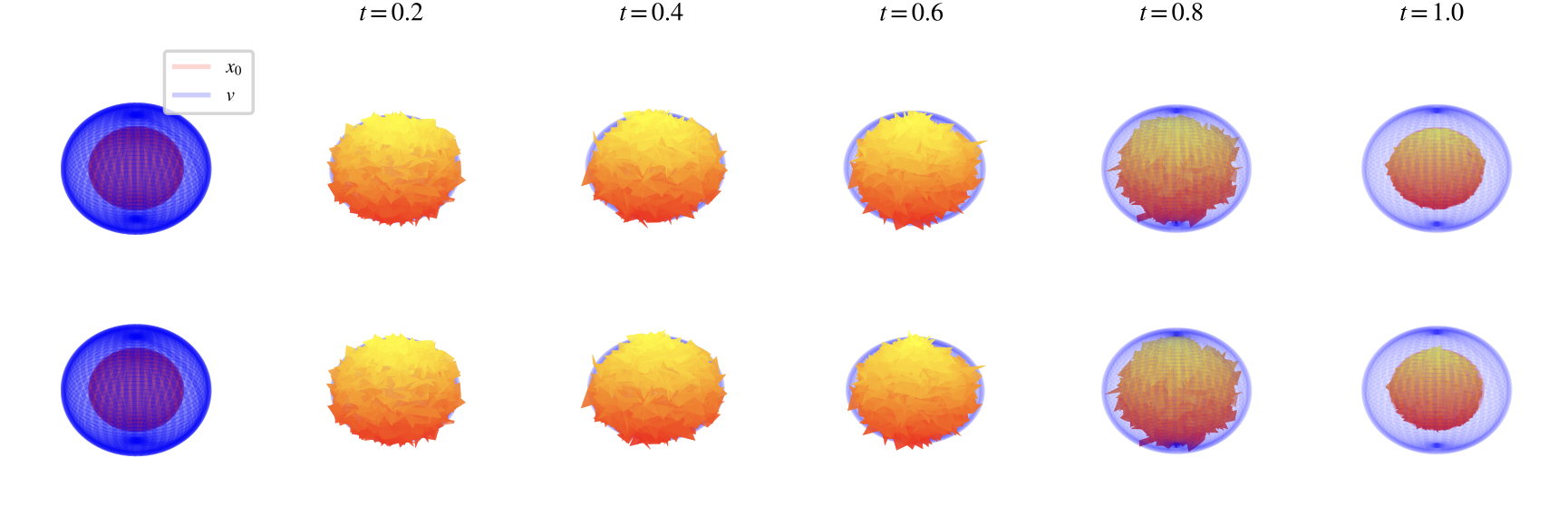}
    \caption{Visualization of the Brownian bridges between two nested spheres with different radii. The top row is the estimated bridge, and the bottom row is the ground true path with the same seed. The intermediate shapes are plotted in light orange. The training is done on \(16\times 16\) grid size and evaluated on \(48\times 48\) grid size.}
    \label{fig: 4}
\end{figure*}

\textbf{Spheres}: Since the construction of our proposed neural operator does not put limitations on the dimension of the domain of the functions, we demonstrate it by modeling the bridges between 3D spheres, where the sphere is considered as a 2D manifold parameterized by a function from \([0,\pi]\times[0, 2\pi]\subset\R^2\) to \(\mathcal{S}^2\subset\R^3\). We discretize the domain evenly into a \(m\times m\) grid and thus represent the sphere function evaluation by a \([m, m, 3]\)-shaped array. During inference, we sample a denser \(m'\times m'\) grid where \(m'>m\), which represents a function evaluation with a higher resolution. Figure \ref{fig: 4} presents a comparison between the estimated bridge and the true bridge, demonstrating an almost indistinguishable difference. Note that the evaluated grid is three times as dense as the training grid, which requires eight times more evaluation points, but our model can still produce a consistent estimation.

\subsection{Nonlinear stochastic landmark matching}
\begin{figure}[!ht]
    \centering
    \begin{subfigure}[b]{0.45\columnwidth}
        \includegraphics[width=\textwidth]{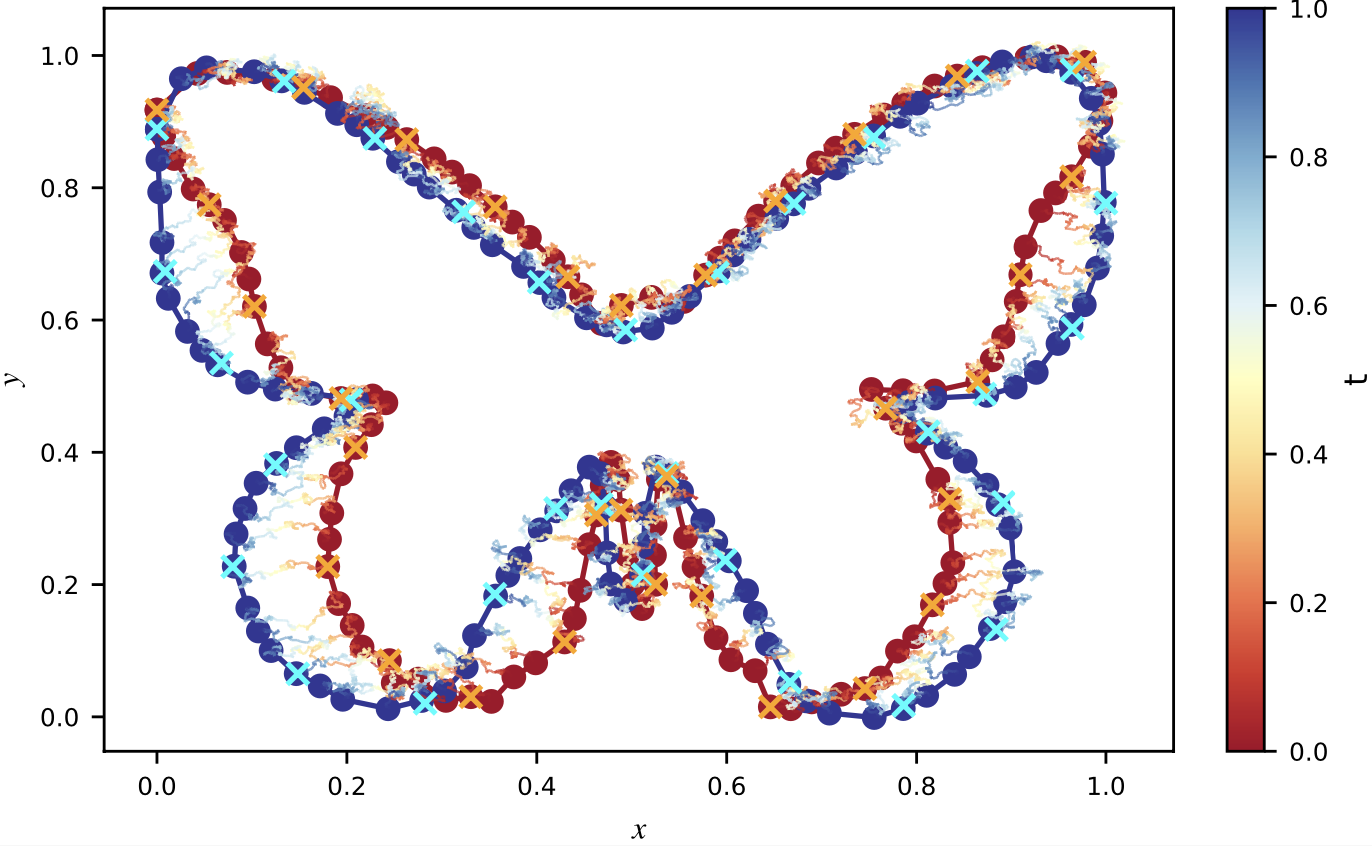}
        \caption{Evaluate on 128 points}
        \label{fig: 5(a)}
    \end{subfigure}
    \begin{subfigure}[b]{0.45\columnwidth}
        \includegraphics[width=\textwidth]{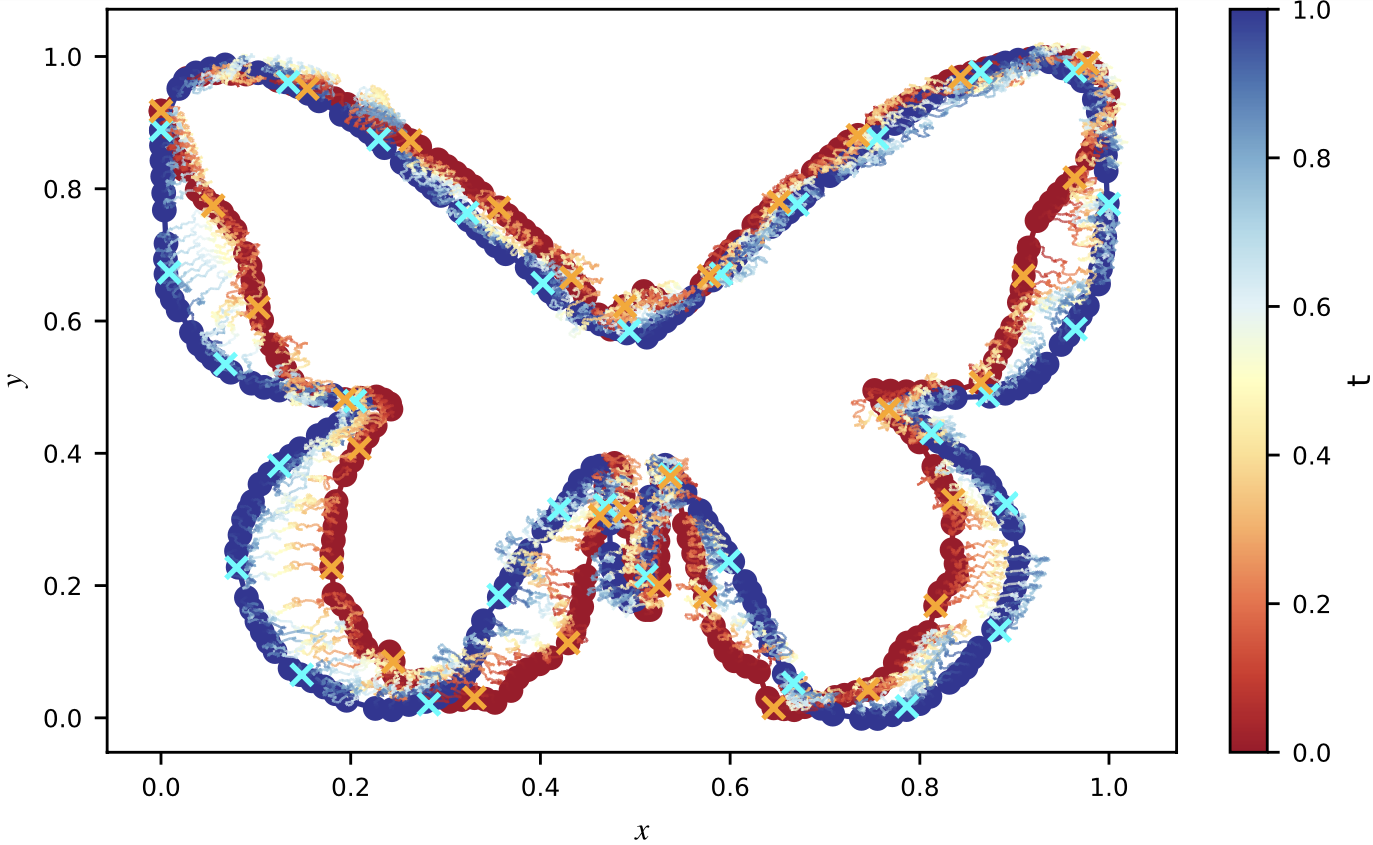}
        \caption{Evaluate on 256 points}
        \label{fig: 5(b)}
    \end{subfigure}

    \caption{Visualization of the nonlinear shape bridges between two butterfly shapes. The cyan crosses mark the landmarks used for training.}
    \label{fig: 5}
\end{figure}

We finally evaluate our method on real data. Specifically, we are interested in modeling the stochastic change in butterfly morphometry characterized by discrete \emph{landmarks} $\{x_i\in\mathbb R^d;\;i=1,\dots,N\}$ over time. In the phylogenetic analysis, such bridge processes model the transitions along the edge between nodes in a phylogenetic tree, where the nodes store the observed data coming from the specimens. Developing the fast bridge simulation approach facilitates large-scale phylogenetic tree simulation and biological inference. The shapes of species are treated as continuous functions $s:\mathcal M\to \mathbb R^d$ for some compact manifold $\mathcal M$, the landmarks are $\{x_i=s(\xi_i);\;\xi_i\in\mathcal M\}$. The change of shapes is realized by applying a time-dependent mapping $\phi_t:\mathbb R^d\to\mathbb R^d$ on $x_i$ to obtain the position $\phi_t(x_i)$ of the $i$-th landmark at time $t$. During the shape evolution, the topology of shape needs to be preserved. Otherwise, the unexpected collapse of landmarks occurs, corresponding to the intersections and overlapping of different parts of the shape. Therefore, $\phi_t$ needs to be diffeomorphic. To model such a process, \cite{sommer2021stochastic, sommer2025stochastic} used a \emph{stochastic flow of landmarks}. Specifically, let $X_t: x\mapsto \phi_t(x)-x$ be an element of the Hilbert space $L^2(\mathbb R^d, \mathbb R^d)$, and $W_t$ be a cylindrical Wiener process on $L^2(\mathbb R^d, \mathbb R^d)$. $X_t$ solves the SDE:
\begin{subequations}
    \begin{align}
        \d X_t &= Q^{1/2}(X_t)\d W_t, \label{eq: kunita_sde}\\
        Q^{1/2}(X_t)(f(\xi)) &= \int_{\R^d}k(X_t(\xi) + \xi,\zeta)f(\zeta)\d \zeta
    \end{align}
\end{subequations}
where the diffusion operator \(Q^{1/2}(X_t)\) on $L^2(\mathbb R^d, \mathbb R^d)$ is a Hilbert-Schmidt, and \(k:\R^d\times\R^d\to\R^d\otimes\mathbb R^d\) is a smooth kernel. It can be shown that \cref{eq: kunita_sde} ensures a diffeomorphism $\phi_t$ for all $t$ \citep{kunita1990stochastic}, where the landmarks are correlated to move synchronically when they are infinitely close, thus, the collapse of landmarks is avoided. We examine our method of modeling the conditional process for \cref{eq: kunita_sde} on real butterfly shapes that are obtained by processing the raw images from gbif.org \citep{gbifbutterflies}. Images are segmented with the Python package \textit{Segment Anything} \citep{kirillov2023segment} and \textit{Gounding Dino} \citep{liu2023grounding}. The assigned landmarks are interpolated to be evenly distributed along the outlines, then aligned by the R package \textit{Geomorph v.4.06} \citep{adams2013geomorph} and normalized. \cref{fig: 5} shows the nonlinear shape bridges between two selected species, Papilio polytes (red) and Parnassius honrathi (blue). We treat the closed outline shape of butterflies as functions \(([0, 1]\subset\R)\to\R^2\). The model is trained on 32 evenly spaced landmarks and evaluated on more landmarks. We can see from the visualization of trajectories that no landmarks collide during the evolution, and the relative positions between landmarks remain. As the number of evaluation points increase, the model can give reasonable interpolations between landmarks. We also test the model on more butterfly species that are close on the phylogenetic tree in \cref{sec: app_butterflies_experiments}, where the simulated trajectory represents the morphological evaluation. 

\section{CONCLUSION}
This paper introduces a novel approach for simulating nonlinear diffusion bridges in infinite-dimensional spaces via operator learning, where we first identify the infinite-dimensional time-reversed diffusion bridge, and propose an objective and time-dependent neural operator architecture to learn the time reversal. We validate our approach through several functional diffusion bridge test cases, where our method showcases its effectiveness of high-resolution zero-shot training and generalization to various shapes, together with consistency under different levels of discretization. Future work will be dedicated to developing a direct forward bridge learning scheme without reversing the bridge and incorporating the endpoints to avoid redundant training for simulating bridges on trees along edges.

\vspace{-5pt}
\section*{Acknowledgements}
\vspace{-5pt}
The authors thank the anonymous reviewers for their valuable comments and suggestions, which have helped improve the clarity and quality of this work. This research was supported by Villum Foundation Grant 40582, the Novo Nordisk Foundation Grant NNF18OC0052000, and Danmarks Frie Forskningsfond Grant 10.46540/3103-00296B.

\bibliographystyle{unsrtnat}
\bibliography{bibfile}

\begin{thebibliography}{44}
\providecommand{\natexlab}[1]{#1}
\providecommand{\url}[1]{\texttt{#1}}
\expandafter\ifx\csname urlstyle\endcsname\relax
  \providecommand{\doi}[1]{doi: #1}\else
  \providecommand{\doi}{doi: \begingroup \urlstyle{rm}\Url}\fi

\bibitem[Shi and Choi(2011)]{shi2011gaussian}
Jian~Qing Shi and Taeryon Choi.
\newblock \emph{Gaussian process regression analysis for functional data}.
\newblock CRC press, 2011.

\bibitem[Rogers and Williams(2000)]{rogers2000diffusions}
Leonard~CG Rogers and David Williams.
\newblock \emph{Diffusions, markov processes, and martingales: Volume 1, foundations}, volume~1.
\newblock Cambridge university press, 2000.

\bibitem[Franzese et~al.(2024)Franzese, Corallo, Rossi, Heinonen, Filippone, and Michiardi]{franzese2024continuous}
Giulio Franzese, Giulio Corallo, Simone Rossi, Markus Heinonen, Maurizio Filippone, and Pietro Michiardi.
\newblock Continuous-time functional diffusion processes.
\newblock \emph{Advances in Neural Information Processing Systems}, 36, 2024.

\bibitem[Lim et~al.(2023)Lim, Kovachki, Baptista, Beckham, Azizzadenesheli, Kossaifi, Voleti, Song, Kreis, Kautz, et~al.]{lim2023score}
Jae~Hyun Lim, Nikola~B Kovachki, Ricardo Baptista, Christopher Beckham, Kamyar Azizzadenesheli, Jean Kossaifi, Vikram Voleti, Jiaming Song, Karsten Kreis, Jan Kautz, et~al.
\newblock Score-based diffusion models in function space.
\newblock \emph{arXiv preprint arXiv:2302.07400}, 2023.

\bibitem[Pidstrigach et~al.(2023)Pidstrigach, Marzouk, Reich, and Wang]{pidstrigach2023infinite}
Jakiw Pidstrigach, Youssef Marzouk, Sebastian Reich, and Sven Wang.
\newblock Infinite-dimensional diffusion models for function spaces.
\newblock \emph{arXiv e-prints}, pages arXiv--2302, 2023.

\bibitem[Baker et~al.(2024)Baker, Yang, Severinsen, Hipsley, and Sommer]{baker2024conditioning}
Elizabeth~Louise Baker, Gefan Yang, Michael~L Severinsen, Christy~Anna Hipsley, and Stefan Sommer.
\newblock Conditioning non-linear and infinite-dimensional diffusion processes.
\newblock \emph{arXiv preprint arXiv:2402.01434}, 2024.

\bibitem[Hyv{\"a}rinen and Dayan(2005)]{hyvarinen2005estimation}
Aapo Hyv{\"a}rinen and Peter Dayan.
\newblock Estimation of non-normalized statistical models by score matching.
\newblock \emph{Journal of Machine Learning Research}, 6\penalty0 (4), 2005.

\bibitem[Vincent(2011)]{vincent2011connection}
Pascal Vincent.
\newblock A connection between score matching and denoising autoencoders.
\newblock \emph{Neural computation}, 23\penalty0 (7):\penalty0 1661--1674, 2011.

\bibitem[Song et~al.(2020)Song, Sohl-Dickstein, Kingma, Kumar, Ermon, and Poole]{song2020score}
Yang Song, Jascha Sohl-Dickstein, Diederik~P Kingma, Abhishek Kumar, Stefano Ermon, and Ben Poole.
\newblock Score-based generative modeling through stochastic differential equations.
\newblock In \emph{International Conference on Learning Representations}, 2020.

\bibitem[Huang et~al.(2021)Huang, Lim, and Courville]{huang2021variational}
Chin-Wei Huang, Jae~Hyun Lim, and Aaron~C Courville.
\newblock A variational perspective on diffusion-based generative models and score matching.
\newblock \emph{Advances in Neural Information Processing Systems}, 34:\penalty0 22863--22876, 2021.

\bibitem[Heng et~al.(2021)Heng, De~Bortoli, Doucet, and Thornton]{heng2021simulating}
Jeremy Heng, Valentin De~Bortoli, Arnaud Doucet, and James Thornton.
\newblock Simulating diffusion bridges with score matching.
\newblock \emph{arXiv preprint arXiv:2111.07243}, 2021.

\bibitem[Clark(1990)]{clark1990simulation}
JMC Clark.
\newblock The simulation of pinned diffusions.
\newblock In \emph{29th IEEE conference on decision and control}, pages 1418--1420. IEEE, 1990.

\bibitem[Delyon and Hu(2006)]{delyon2006simulation}
Bernard Delyon and Ying Hu.
\newblock Simulation of conditioned diffusion and application to parameter estimation.
\newblock \emph{Stochastic Processes and their Applications}, 116\penalty0 (11):\penalty0 1660--1675, 2006.

\bibitem[Schauer et~al.(2017)Schauer, Van Der~Meulen, and Van~Zanten]{schauer2017guided}
Moritz Schauer, Frank Van Der~Meulen, and Harry Van~Zanten.
\newblock Guided proposals for simulating multi-dimensional diffusion bridges, 2017.

\bibitem[Mider et~al.(2021)Mider, Schauer, and Van~der Meulen]{mider2021continuous}
Marcin Mider, Moritz Schauer, and Frank Van~der Meulen.
\newblock Continuous-discrete smoothing of diffusions.
\newblock \emph{Electronic Journal of Statistics}, 15\penalty0 (2):\penalty0 4295--4342, 2021.

\bibitem[Chau et~al.(2024)Chau, Kirkby, Nguyen, Nguyen, Nguyen, and Nguyen]{chau2024efficient}
H.~Chau, J.~L. Kirkby, D.~H. Nguyen, D.~Nguyen, N.~Nguyen, and T.~Nguyen.
\newblock An efficient method to simulate diffusion bridges.
\newblock \emph{Statistics and Computing}, 34\penalty0 (4):\penalty0 131, August 2024.
\newblock ISSN 0960-3174, 1573-1375.
\newblock \doi{10.1007/s11222-024-10439-z}.
\newblock URL \url{https://link.springer.com/10.1007/s11222-024-10439-z}.

\bibitem[Sohl-Dickstein et~al.(2015)Sohl-Dickstein, Weiss, Maheswaranathan, and Ganguli]{sohl2015deep}
Jascha Sohl-Dickstein, Eric Weiss, Niru Maheswaranathan, and Surya Ganguli.
\newblock Deep unsupervised learning using nonequilibrium thermodynamics.
\newblock In \emph{International conference on machine learning}, pages 2256--2265. PMLR, 2015.

\bibitem[Song et~al.(2021)Song, Sohl-Dickstein, Kingma, Kumar, Ermon, and Poole]{song2021scorebased}
Yang Song, Jascha Sohl-Dickstein, Diederik~P Kingma, Abhishek Kumar, Stefano Ermon, and Ben Poole.
\newblock Score-based generative modeling through stochastic differential equations.
\newblock In \emph{International Conference on Learning Representations}, 2021.
\newblock URL \url{https://openreview.net/forum?id=PxTIG12RRHS}.

\bibitem[Baldassari et~al.(2024)Baldassari, Siahkoohi, Garnier, Solna, and de~Hoop]{baldassari2024conditional}
Lorenzo Baldassari, Ali Siahkoohi, Josselin Garnier, Knut Solna, and Maarten~V de~Hoop.
\newblock Conditional score-based diffusion models for bayesian inference in infinite dimensions.
\newblock \emph{Advances in Neural Information Processing Systems}, 36, 2024.

\bibitem[Hagemann et~al.(2023)Hagemann, Mildenberger, Ruthotto, Steidl, and Yang]{hagemann2023multilevel}
Paul Hagemann, Sophie Mildenberger, Lars Ruthotto, Gabriele Steidl, and Nicole~Tianjiao Yang.
\newblock Multilevel diffusion: Infinite dimensional score-based diffusion models for image generation.
\newblock \emph{arXiv preprint arXiv:2303.04772}, 2023.

\bibitem[De~Bortoli et~al.(2021)De~Bortoli, Thornton, Heng, and Doucet]{de2021diffusion}
Valentin De~Bortoli, James Thornton, Jeremy Heng, and Arnaud Doucet.
\newblock Diffusion schrödinger bridge with applications to score-based generative modeling.
\newblock \emph{Advances in Neural Information Processing Systems}, 34:\penalty0 17695--17709, 2021.

\bibitem[Shi et~al.(2024)Shi, De~Bortoli, Campbell, and Doucet]{shi2024diffusion}
Yuyang Shi, Valentin De~Bortoli, Andrew Campbell, and Arnaud Doucet.
\newblock Diffusion schrödinger bridge matching.
\newblock \emph{Advances in Neural Information Processing Systems}, 36, 2024.

\bibitem[Tang et~al.(2024)Tang, Hang, Gu, Chen, and Guo]{tang2024simplified}
Zhicong Tang, Tiankai Hang, Shuyang Gu, Dong Chen, and Baining Guo.
\newblock Simplified diffusion schrödinger bridge.
\newblock \emph{arXiv preprint arXiv:2403.14623}, 2024.

\bibitem[Thornton et~al.(2022)Thornton, Hutchinson, Mathieu, De~Bortoli, Teh, and Doucet]{thornton2022riemannian}
James Thornton, Michael Hutchinson, Emile Mathieu, Valentin De~Bortoli, Yee~Whye Teh, and Arnaud Doucet.
\newblock Riemannian diffusion schrödinger bridge.
\newblock \emph{arXiv preprint arXiv:2207.03024}, 2022.

\bibitem[Haussmann and Pardoux(1986)]{haussmann1986time}
Ulrich~G Haussmann and Etienne Pardoux.
\newblock Time reversal of diffusions.
\newblock \emph{The Annals of Probability}, pages 1188--1205, 1986.

\bibitem[Millet et~al.(1989)Millet, Nualart, and Sanz]{millet1989time}
Annie Millet, David Nualart, and Marta Sanz.
\newblock Time reversal for infinite-dimensional diffusions.
\newblock \emph{Probability theory and related fields}, 82\penalty0 (3):\penalty0 315--347, 1989.

\bibitem[Saharia et~al.(2022)Saharia, Chan, Saxena, Li, Whang, Denton, Ghasemipour, Gontijo~Lopes, Karagol~Ayan, Salimans, et~al.]{saharia2022photorealistic}
Chitwan Saharia, William Chan, Saurabh Saxena, Lala Li, Jay Whang, Emily~L Denton, Kamyar Ghasemipour, Raphael Gontijo~Lopes, Burcu Karagol~Ayan, Tim Salimans, et~al.
\newblock Photorealistic text-to-image diffusion models with deep language understanding.
\newblock \emph{Advances in neural information processing systems}, 35:\penalty0 36479--36494, 2022.

\bibitem[Vaswani et~al.(2017)Vaswani, Shazeer, Parmar, Uszkoreit, Jones, Gomez, Kaiser, and Polosukhin]{vaswani2017attention}
Ashish Vaswani, Noam Shazeer, Niki Parmar, Jakob Uszkoreit, Llion Jones, Aidan~N Gomez, Lukasz Kaiser, and Illia Polosukhin.
\newblock Attention is all you need.
\newblock \emph{Advances in neural information processing systems}, 30, 2017.

\bibitem[Li et~al.(2020{\natexlab{a}})Li, Kovachki, Azizzadenesheli, Bhattacharya, Stuart, Anandkumar, et~al.]{li2020fourier}
Zongyi Li, Nikola~Borislavov Kovachki, Kamyar Azizzadenesheli, Kaushik Bhattacharya, Andrew Stuart, Anima Anandkumar, et~al.
\newblock Fourier neural operator for parametric partial differential equations.
\newblock In \emph{International Conference on Learning Representations}, 2020{\natexlab{a}}.

\bibitem[Li et~al.(2020{\natexlab{b}})Li, Kovachki, Azizzadenesheli, Liu, Bhattacharya, Stuart, and Anandkumar]{li2020neural}
Zongyi Li, Nikola Kovachki, Kamyar Azizzadenesheli, Burigede Liu, Kaushik Bhattacharya, Andrew Stuart, and Anima Anandkumar.
\newblock Neural operator: Graph kernel network for partial differential equations.
\newblock \emph{arXiv preprint arXiv:2003.03485}, 2020{\natexlab{b}}.

\bibitem[Lu et~al.(2021)Lu, Jin, Pang, Zhang, and Karniadakis]{lu2021learning}
Lu~Lu, Pengzhan Jin, Guofei Pang, Zhongqiang Zhang, and George~Em Karniadakis.
\newblock Learning nonlinear operators via deeponet based on the universal approximation theorem of operators.
\newblock \emph{Nature machine intelligence}, 3\penalty0 (3):\penalty0 218--229, 2021.

\bibitem[Kovachki et~al.(2023)Kovachki, Li, Liu, Azizzadenesheli, Bhattacharya, Stuart, and Anandkumar]{kovachki2023neural}
Nikola Kovachki, Zongyi Li, Burigede Liu, Kamyar Azizzadenesheli, Kaushik Bhattacharya, Andrew Stuart, and Anima Anandkumar.
\newblock Neural operator: Learning maps between function spaces with applications to pdes.
\newblock \emph{Journal of Machine Learning Research}, 24\penalty0 (89):\penalty0 1--97, 2023.

\bibitem[Park et~al.(2023)Park, Choi, Yoon, Kang, et~al.]{park2023learning}
Yesom Park, Jaemoo Choi, Changyeon Yoon, Myungjoo Kang, et~al.
\newblock Learning pde solution operator for continuous modeling of time-series.
\newblock \emph{arXiv preprint arXiv:2302.00854}, 2023.

\bibitem[Ashiqur~Rahman et~al.(2022)Ashiqur~Rahman, Ross, and Azizzadenesheli]{ashiqur2022u}
Md~Ashiqur~Rahman, Zachary~E Ross, and Kamyar Azizzadenesheli.
\newblock U-no: U-shaped neural operators.
\newblock \emph{arXiv e-prints}, pages arXiv--2204, 2022.

\bibitem[Da~Prato and Zabczyk(2014)]{da2014stochastic}
Giuseppe Da~Prato and Jerzy Zabczyk.
\newblock \emph{Stochastic equations in infinite dimensions}.
\newblock Cambridge university press, 2014.

\bibitem[Phillips et~al.(2022)Phillips, Seror, Hutchinson, De~Bortoli, Doucet, and Mathieu]{phillips2022spectral}
Angus Phillips, Thomas Seror, Michael Hutchinson, Valentin De~Bortoli, Arnaud Doucet, and Emile Mathieu.
\newblock Spectral diffusion processes.
\newblock \emph{arXiv preprint arXiv:2209.14125}, 2022.

\bibitem[Sommer et~al.(2021)Sommer, Schauer, and Meulen]{sommer2021stochastic}
{Stefan Horst} Sommer, Moritz Schauer, and {Frank van der} Meulen.
\newblock Stochastic flows and shape bridges.
\newblock In \emph{Statistics of Stochastic Differential Equations on Manifolds and Stratified Spaces (hybrid meeting)}, number~48 in Oberwolfach Reports, pages 18--21. Mathematisches Forschungsinstitut Oberwolfach, 2021.
\newblock \doi{10.4171/OWR/2021/48}.
\newblock null ; Conference date: 03-10-2021 Through 09-10-2021.

\bibitem[Sommer et~al.(2025)Sommer, Yang, and Louise]{sommer2025stochastic}
Stefan Sommer, Gefan Yang, and Baker~Elizabeth Louise.
\newblock Stochastics of shapes and kunita flows.
\newblock \emph{Submitted}, 2025.

\bibitem[Kunita(1990)]{kunita1990stochastic}
Hiroshi Kunita.
\newblock \emph{Stochastic flows and stochastic differential equations}, volume~24.
\newblock Cambridge university press, 1990.

\bibitem[{GBIF.Org User}(2024)]{gbifbutterflies}
{GBIF.Org User}.
\newblock Occurrence download, 2024.
\newblock URL \url{https://www.gbif.org/occurrence/download/0075323-231120084113126}.

\bibitem[Kirillov et~al.(2023)Kirillov, Mintun, Ravi, Mao, Rolland, Gustafson, Xiao, Whitehead, Berg, Lo, et~al.]{kirillov2023segment}
Alexander Kirillov, Eric Mintun, Nikhila Ravi, Hanzi Mao, Chloe Rolland, Laura Gustafson, Tete Xiao, Spencer Whitehead, Alexander~C Berg, Wan-Yen Lo, et~al.
\newblock Segment anything.
\newblock In \emph{Proceedings of the IEEE/CVF International Conference on Computer Vision}, pages 4015--4026, 2023.

\bibitem[Liu et~al.(2023)Liu, Zeng, Ren, Li, Zhang, Yang, Li, Yang, Su, Zhu, et~al.]{liu2023grounding}
Shilong Liu, Zhaoyang Zeng, Tianhe Ren, Feng Li, Hao Zhang, Jie Yang, Chunyuan Li, Jianwei Yang, Hang Su, Jun Zhu, et~al.
\newblock Grounding dino: Marrying dino with grounded pre-training for open-set object detection.
\newblock \emph{arXiv preprint arXiv:2303.05499}, 2023.

\bibitem[Adams and Ot{\'a}rola-Castillo(2013)]{adams2013geomorph}
Dean~C Adams and Erik Ot{\'a}rola-Castillo.
\newblock geomorph: an r package for the collection and analysis of geometric morphometric shape data.
\newblock \emph{Methods in ecology and evolution}, 4\penalty0 (4):\penalty0 393--399, 2013.

\bibitem[Heek et~al.(2024)Heek, Levskaya, Oliver, Ritter, Rondepierre, Steiner, and van {Z}ee]{flax2020github}
Jonathan Heek, Anselm Levskaya, Avital Oliver, Marvin Ritter, Bertrand Rondepierre, Andreas Steiner, and Marc van {Z}ee.
\newblock {F}lax: A neural network library and ecosystem for {JAX}, 2024.
\newblock URL \url{http://github.com/google/flax}.

\end{thebibliography}

\onecolumn
\aistatstitle{Infinite-dimensional Diffusion Bridge Simulation via Operator Learning: Supplementary Materials}

\makeatletter
\renewcommand{\section}{\@startsection{section}{1}{0pt}{-3.5ex plus -1ex minus -.2ex}{2.3ex plus .2ex}{\normalfont\Large\bfseries}}
\makeatother

\section{PROOFS}
In this section, we detail the proofs of the two theorems and one lemma stated in the paper.

\subsection{Proof of \texorpdfstring{\cref{thm: t_reversal}}{Theorem 2}} \label{sec: t_reversal_thm_proof}

\begin{proof}
    We follow the setup in \citep{millet1989time}. Let \(\{e_i\}_{i\in\mathbb Z}\) be an orthonormal basis for \(\H\ni \Xs_t\), and \(\{k_j\}_{j\in\mathbb Z}\) be an orthonormal basis for \(\U\ni W_t\). Then suppose that for each \(i\), there exists a finite set \(I(i)\), such that for all \(s\), \([a(s, x)]_{ij}\coloneq\sum_\ell^{\infty}[g(s,x)]_{i\ell}[g(s,x)]_{j\ell}=0\) if \(j\notin I(i)\). Then we split \(\Xs_t\) into an infinite number of finite vectors; specifically, we denote \(\bxs{i}_t=\{[\Xs_t]_j;j\in I(i)\}\in\R^{|I(i)|}\) to be the vector consisting of all the components of \(\Xs_t\) indexed by the entries in \(I(i)\), with \([\Xs_t]_i = \langle \Xs_t, e_j \rangle_{\H}\), and its complement to be \(\hbxs{i}_t=\{[\Xs_t]_j;j\notin I(i)\}\), and let \(\bzs{i}=\{z_j\in\R;\;j\notin I(i)\}\). Therefore, \(\Xs_t=\{\bxs{i}_t, \hbxs{i}_t\}\). We assume that the conditional law of \(\bxs{i}_t\) given \(\hbxs{i}_t=\bzs{i}\) has a density \(\tilde{p}(\bxs{i}_t\mid\bzs{i})\) with respect to the Lebesgue measure in the vector space spanned by \(\{e_j;\; j\in I(i)\}\). We refer to \citep[Section 5]{millet1989time} for the existence of such a density.
    Consider the infinite-dimensional bridge process \(\Xs\) defined by \cref{eq: inf_doob_sde}, written in terms of the bases \(e_i\) and \(k_j\):
    \begin{equation}
        \d \Xs_t = [f(t, \Xs_t) + a(t, \Xs_t)\nabla\log h(t, \Xs_t)]_i\d t + \sum^{\infty}_{j=1}[g(t, \Xs_t)]_{ij}\d [W^{\Ps}_t]_j.
    \end{equation}
    Then apply the time reversal theorem \citep[Theorem 4.3]{millet1989time} to obtain the reversed bridge \(\Ys\) in terms of basis coefficients:
    \begin{equation}
        \d \Ys_t = [\overline{f^{\star}}(t, \Ys_t)]_i \d t + \sum^{\infty}_{j=1}[\overline{g^{\star}}(t, \Ys_t)]_{ij}\d [W^{\Ps}_t]_j,
    \end{equation}
    where,
    \begin{subequations}
        \begin{align}
            [\overline{f^{\star}}(s, x)]_i &= -[f(T-s, x) + a(T-s, x)\nabla\log h(T-s, x)]_i \nonumber\\
            &\quad\: + \tilde{p}^{-1}(\bxs{i}\mid\bzs{i})\sum_{j\in I(i)}\nabla_{[x]_j}\{[a(T-s, x)]_{ij}\tilde{p}(\bxs{i}|\bzs{i})\}, \\
            [\overline{g^{\star}}(s, x)]_{ij} &= [g(T-s, x)]_{ij}.
        \end{align}
    \end{subequations}
    Using \(p^{-1}\nabla p=\nabla\log p\), we can rewrite \([\overline{f^{\star}}(s, x)]_i\) as
    \begin{align}
        [\overline{f^{\star}}(s, x)]_i = &-[f(T-s, x) + a(T-s, x)\nabla\log h(T-s, x)]_i \nonumber\\
        &+ \sum_{j\in I(i)}\nabla_{[x]_j}[a(T-s, x)]_{ij} + \sum_{j\in I(i)}[a(T-s, x)]_{ij}\nabla_{[x]_j} \log \tilde{p}(\bxs{i}|\bzs{i}).
    \end{align}
    We then focus on the sum
    \begin{equation}    \label{eq: sum}
        -[a(T-s, x)\nabla\log h(T-s, x)]_i + \sum_{j\in I(i)}[a(T-s, x)]_{ij}\nabla_{[x]_j}\log \tilde{p}(\bxs{i}|\bzs{i}).
    \end{equation}
    Noting that \([a(t, x)]_{ij}=0\) for any \(j\notin I(i)\), we expand the first term into basis elements:
    \begin{equation}
        [a(T-s, x)\nabla\log h(T-s, x)]_i = \sum_{j\in I(i)}[a(T-s, x)]_{ij}\nabla_{[x]_j}\log h(T-s, x).
    \end{equation}
    Hence, \cref{eq: sum} becomes
    \begin{equation}
        \sum_{j\in I(i)}[a(T-t, x)]_{ij}\nabla_{[x]_j}\log\left[\frac{\tilde{p}(\bxs{i}\mid\bzs{i})}{h(T-t, x)}\right].
    \end{equation}
    Finally, we note that \(\tilde{p}(\bxs{i}\mid\bzs{i})\) is the \(h\)-transformation of \(p(\bx{i}\mid (\bz{i}, x_0))\), where \(p(\bx{i}\mid(\bz{i}, x_0))\) is the conditional law of \(\bx{i}\) given \(\hbx{i}=\bz{i}\) and \(X_0=x_0\). Hence,
    \begin{equation}
        \tilde{p}(\bxs{i}\mid\bzs{i}) = p(\bx{i}\mid(\bz{i}, x_0))h(T-s, x).
    \end{equation}
    Substituting it back gives the desired form stated in the theorem
    \begin{equation}
        [\bar{f}(s, x)]_i = -[f(T-s, x)]_i + \sum_{j\in I(i)}\nabla_{[x]_j}[a(T-s,x)]_{ij} + \sum_{j\in I(i)}[a(T-s, x)]_{ij}\nabla_{[x]_j} \log p(\bx{i}\mid(\bz{i}, x_0)).
    \end{equation}
\end{proof}

\subsection{Proof of \texorpdfstring{\cref{lem: kld}}{Lemma 1}}
\label{sec: kld_lemma_proof}
\begin{proof}
    By definition, the KL divergence between \(\P\) and \(\Pt\) can be expressed as the expectation:
    \begin{equation}
        \kld(\P||\Pt) = \E_{\P}\left[\frac{\d \L^{\P}}{\d \L^{\Pt}}(Y)\right] =\E_{\P}\left[\log\frac{\d \P}{\d \Pt}\right],
    \end{equation}
    where \(\L^{\P}, \L^{\Pt}\) denote the laws of \(Y\) under \(\P, \Pt\) respectively. With a change of variable notation from \(y\) to \(x\) with reversing the time direction from \(T-s\) to \(s\) , the Radon-Nikodym derivative \(\frac{\d \P}{\d \Pt}\) can be computed by Girsanov's theorem (see e.g., \cite{da2014stochastic}) as:
    \begin{equation}
        \frac{\d \P}{\d \Pt} = \exp\left\{\int^{T}_{0}\left\{g^{*}(\bar{f} - f^{(\theta)})\right\}(s, x)\,\d W^{\P}_s + \frac{1}{2}\int^{T}_{0}\left\|\left\{g^{*}(\bar{f} - f^{(\theta)})\right\}(s, x)\right\|^2_{Q^{1/2}(\U)}\,\d s\right\}.
    \end{equation}
    The first Itô integral is a zero-mean martingale, therefore, it immediately reads:
    \begin{equation}
         \kld(\P||\Pt) = \frac{1}{2}\E_{\P}\left[ \int^{T}_{0}\left\|\left\{g^{*}(\bar{f} - f^{(\theta)})\right\}(s, x)\right\|^2_{Q^{1/2}(\U)}\,\d s\right],
    \end{equation}
    where \(\|\cdot\|_{Q^{1/2}(\U)}\) denotes the norm in \(Q^{1/2}(\U)\). Introducing the basis element \(g^*(s, x)e_i\in Q^{1/2}(\U)\) and using the equivalence \(\langle g^* e_i, g^* e_j\rangle_{Q^{1/2}(\U)}=\langle gg^* e_i, e_j\rangle_{\H}\), we see:
    \begin{subequations}
        \begin{align}
        \kld(\P||\Pt) &= \frac{1}{2}\E_{\P}\left[ \int^{T}_{0}\sum^{\infty}_{i=1}\left\langle \left\{g^*(\bar{f}-f^{(\theta)})\right\}(s, x), g^*(s, x)e_i\right\rangle^2_{Q^{1/2}(\U)}\,\d s\right] \\
        &= \frac{1}{2}\E_{\P}\left[ \int^{T}_{0}\sum^{\infty}_{i=1}\left\langle \{\bar{f}-f^{(\theta)}\}(s, x), gg^*(s, x)e_i\right\rangle^2_{\H}\,\d s\right] \\
        &= \frac{1}{2}\E_{\P}\left[ \int^{T}_{0}\sum^{\infty}_{i=1} \lambda_i [\{\bar{f}-f\}(s, x)]^2_i \,\d s\right] \\
        &= \frac{1}{2}\E_{\P}\left[ \int^{T}_{0}\sum^{\infty}_{i=1} \lambda_i [\Psi(s, x)]^2_i \,\d s\right].
        \end{align}
    \end{subequations}
    The second last equivalence comes from \(a(s, x)\) being diagonal in \(\{e_i\}\), i.e., \(a(e_i)=\lambda_i e_i\) for some scalar \(\lambda_i\). To derive \(\Psi(s, x)\), consider the definitions of \(\bar{f}\) and \(f^{(\theta)}\):
    \begin{subequations}
        \begin{align}
            [\bar{f}(T-s, x)]_i &= -[f(s, x)]_i + p^{-1}(\bx{i}|\bz{i})\sum_{j\in I(i)}\nabla_j\left([a(s, x)]_{ij}p(\bx{i}| \bz{i})\right) \label{eq: f_bar}\\
            [f^{(\theta)}(T-s, x)]_i &= -[f(s, x)]_i + \sum_{j\in I(i)}\nabla_{[x]_j}[a(s, x)]_{ij} + [\mathcal{G}^{(\theta)}(s, x)]_i.
        \end{align}
    \end{subequations}
    We focus on the product in the RHS of \cref{eq: f_bar}:
    \begin{subequations}
        \begin{align}
            &p^{-1}(\bx{i}\mid\bz{i})\sum_{j\in I(i)}\nabla_j\left([a(s, x)]_{ij}p(\bx{i}\mid \bz{i})\right) \\
            &= p^{-1}(\bx{i}\mid\bz{i})\left\{\sum_{j\in I(i)}p(\bx{i}\mid\bz{i})\nabla_{j}[a(s, x)]_{ij} + \sum_{j\in I(i)}[a(s, x)]_{ij}\nabla_{j}p(\bx{i}\mid\bz{i})\right\} \\
            &=\sum_{j\in I(i)}\nabla_{j}[a(s, x)]_{ij} + \sum_{j\in I(i)}[a(s, x)]_{ij}\nabla_{j}\log p(\bx{i}\mid\bz{i}).
        \end{align}
    \end{subequations}
    Then it turns out that:
    \begin{equation}
        \Psi(s, x) = \sum_{j\in I(i)}[a(s, x)]_{ij}\nabla_{j}\log p(\bx{i}\,|\,\bz{i}) - [\mathcal{G}^{(\theta)}(s, x)]_i
    \end{equation}
\end{proof}

\subsection{Proof of \texorpdfstring{\cref{thm: loss}}{Theorem 3}} \label{sec: loss_thm_proof}

\begin{proof}
    We follow the derived expression of \(\kld(\P||\Pt)\) in \cref{lem: kld}, and denote \(\nabla_j := \nabla_{[x]_j}=\partial /\partial [x]_j\) for simplicity:
    \begin{subequations}
        \begin{align}
            \kld(\P||\Pt) 
            &= \frac{1}{2}\E_{\P}\left[\int^{T}_0\sum^{\infty}_{i=1}\lambda_i\left\{\sum_{j\in I(i)}[a(s, x)]_{ij}\nabla_{j}\log p(\bx{i}|\bz{i}) - [\mathcal{G}^{(\theta)}(s, x)]_i\right\}^2\d t\right] \\
            &= \underbrace{\frac{1}{2}\E_{\P}\left[ \int^{T}_{0}\sum^{\infty}_{i=1}\lambda_i[\mathcal{G}^{(\theta)}(t, x)]_i^2\d t\right]}_{C_1} \nonumber \\
            &\quad + \underbrace{\frac{1}{2}\E_{\P}\left[ \int^{T}_{0}\sum^{\infty}_{i=1}\lambda_i\left\{ \sum_{j\in I(i)}[a(t, x)]_{ij}\nabla_j\log p(\bx{i}|\bz{i})\right\}^2\,\d t\right]}_{C_2} \nonumber \\
            &\quad -\underbrace{\E_{\P}\left[\int^{T}_{0}\sum^{\infty}_{i=1}\lambda_i\left\langle \sum_{j\in I(i)}[a(t, x)]_{ij}\nabla_j\log p(\bx{i}|\bz{i}), [\mathcal{G}^{(\theta)}(t, x)]_i\right\rangle\d t\right]}_{C_3}.
        \end{align}
    \end{subequations}
    We then focus on \(C_3\): for any time partition \((t_n)^{N}_{n=1}\), we define \(\bx{i}_{t_n}, \hbx{i}_{t_n}\), and \(\bz{i}_{t_n}\) for \(X_{t_n}\) in the same manner as \(\bx{i}, \hbx{i}\), and \(\bz{i}\). Then we can write \(C_3\) as
    \begin{subequations}
        \begin{align}
            C_3 &= \E_{\P}\left[\sum_{n=1}^{N}\int^{t_n}_{t_{n-1}}\sum^{\infty}_{i=1}\lambda_i\left\langle \sum_{j\in I(i)}[a(t, x)]_{ij}\nabla_j\log p(\bx{i}|\bz{i}), [\mathcal{G}^{(\theta)}(t, x)]_i\right\rangle\d t\right] \\
            &= \sum_{n=1}^{N}\sum^{\infty}_{i=1}\lambda_i\int_{\Omega=\mathrm{span}(\{e_j;j\in I(i)\})}p(\bx{i} \mid \bz{i})\int^{t_n}_{t_{n-1}}\left\langle \sum_{j\in I(i)}[a(t, x)]_{ij}\nabla_j\log p_t(\bx{i} \mid \bz{i}), [\mathcal{G}^{(\theta)}(t, x)]_i\right\rangle \d\bx{i}\d t \\
            &= \sum_{n=1}^{N}\sum^{\infty}_{i=1}\lambda_i\int_{\Omega}\int^{t_n}_{t_{n-1}}\left\langle \sum_{j\in I(i)}[a(t, x)]_{ij}\nabla_j p(\bx{i}\mid \bz{i}), [\mathcal{G}^{(\theta)}(t, x)]_i\right\rangle \d\bx{i}\d t .\label{eq: c3_inner}
        \end{align}
    \end{subequations}
    By using the Chapman-Kolmogorov equation \citep[Corollary 9.15]{da2014stochastic}, one can write \(p_t(\bx{i}|\bz{i})\) as
    \begin{equation}    \label{eq: ck_eq}
        p(\bx{i}|\bz{i}) = \int_\Omega p(\bx{i}_0|\bz{i}_0)\d \bx{i}_0 \int_\Omega p(\bx{i}\mid (\bx{i}_{t_{n-1}},\bz{i},\bz{i}_{t_{n-1}}))
        p(\bx{i}_{t_{n-1}}\mid(\bz{i}_{t_{n-1}}, \bx{i}_0, \bz{i}_0))\d\bx{i}_{t_{n-1}}.
    \end{equation} 
    We can then differentiate \cref{eq: ck_eq}:
    \begin{subequations}
        \begin{align}
            &\nabla_j p(\bx{i}|\bz{i}) \nonumber \\
            &= \int_\Omega p(\bx{i}_0\mid\bz{i}_0)\d \bx{i}_0 \int_\Omega \nabla_j p(\bx{i}\mid(\bx{i}_{t_{n-1}},\bz{i},\bz{i}_{t_{n-1}}))p(\bx{i}_{t_{n-1}}\mid(\bz{i}_{t_{n-1}}, \bx{i}_0, \bz{i}_0))\d\bx{i}_{t_{n-1}} \\
            &= \int_\Omega p(\bx{i}_0\mid\bz{i}_0)\d \bx{i}_0 \int_\Omega \nabla_j \log p(\bx{i}\mid(\bx{i}_{t_{n-1}},\bz{i},\bz{i}_{t_{n-1}})) \nonumber\\
            &\qquad p(\bx{i}\mid(\bx{i}_{t_{n-1}},\bz{i},\bz{i}_{t_{n-1}}))p(\bx{i}_{t_{n-1}}\mid(\bz{i}_{t_{n-1}}, \bx{i}_0, \bz{i}_0))\d\bx{i}_{t_{n-1}} \\
            &= \int_\Omega p(\bx{i}_0\mid\bz{i}_0)\d \bx{i}_0 \int_\Omega \nabla_j \log p(\bx{i}\mid(\bx{i}_{t_{n-1}},\bz{i},\bz{i}_{t_{n-1}})) \nonumber \\
            &\qquad p(\bx{i}\mid(\bx{i}_{t_{n-1}},\bz{i},\bz{i}_{t_{n-1}}))p(\bx{i}_{t_{n-1}}\mid(\bz{i}_{t_{n-1}}, \bx{i}_0, \bz{i}_0))\d\bx{i}_{t_{n-1}}.
        \end{align}
    \end{subequations}
    Substituting it back into \cref{eq: c3_inner} gives
    \begin{subequations}
        \begin{align}
            C_3 &= \sum_{n=1}^{N}\sum^{\infty}_{i=1}\lambda_i\int_{\Omega}p_t(\bx{i}|\bz{i}) \nonumber \\
            &\qquad \int^{t_n}_{t_{n-1}}\left\langle \sum_{j\in I(i)}[a(t, x)]_{ij}\nabla_j \log p(\bx{i}\mid(\bz{i}, \bx{i}_{t_{n-1}}, \bz{i}_{t_{n-1}})), [\mathcal{G}^{(\theta)}(t, x)]_i\right\rangle \d\bx{i}\d t \\
            &= \E_{\P}\left[ \sum_{n=1}^{N}\sum^{\infty}_{i=1}\lambda_i\left\langle \sum_{j\in I(i)}[a(t, x)]_{ij}\nabla_j \log p(\bx{i}\mid(\bz{i}, X_{t_{n-1}})), [\mathcal{G}^{(\theta)}(t, x)]_i\right\rangle \d t\right].
        \end{align}
    \end{subequations}
    We then claim that:
    \begin{equation}    \label{eq: claim}
        L(\theta) = \frac{1}{2}\sum^{N}_{n=1}\int^{t_n}_{t_{n-1}}\mathbb{E}_{\P}\left[ \sum_{i=1}^{\infty}\lambda_i\left\{ [\mathcal{G}^{(\theta)}(t, X_t)]_i - [b_t(X_t, x_{t_{n-1}})]_i \right\}^2\right]\d t = C_1 + C_4 - C_3,
    \end{equation}
    where,
    \begin{equation}
        C_4 = \frac{1}{2}\E_{\P}\left[\sum^N_{n=1}\int^{t_{n}}_{t_{n-1}}\sum^{\infty}_{i=1}\lambda_i\left\{\sum_{j\in I(i)}[a(t, x)]_{ij}\nabla_j\log p(\bx{i}\mid(\bz{i}, x_{t_{n-1}}))\right\}^2\d t\right].
    \end{equation}
    To show this, we expand \cref{eq: claim}:
    \begin{align}
        L(\theta) &= \underbrace{\frac{1}{2}\E_{\P}\left[ \sum^{N}_{n=1} \int^{t_n}_{t_{n-1}}\sum^{\infty}_{i=1}\lambda_i[\mathcal{G}^{(\theta)}(t, x)]_i^2\d t\right]}_{C_1} \nonumber \\
        &\quad + \underbrace{\frac{1}{2}\E_{\P}\left[ \sum^{N}_{n=1} \int^{t_n}_{t_{n-1}}\sum^{\infty}_{i=1}\lambda_i\left\{ \sum_{j\in I(i)}[a(t, x)]_{ij}\nabla_j\log p(\bx{i}\mid(\bz{i}, x_{t_{n-1}}))\right\}^2\d t\right]}_{C_4} \nonumber \\
        &\quad - \underbrace{\E_{\Ps}\left[\sum^{N}_{n=1} \int^{t_n}_{t_{n-1}}\sum^{\infty}_{i=1}\lambda_i\left\langle \sum_{j\in I(i)}[a(t, x)]_{ij}\nabla_j\log p(\bx{i}\mid (\bz{i}, x_{t_{n-1}})), [\mathcal{G}^{(\theta)}(t, x)]_i\right\rangle\d t\right]}_{C_3}.
    \end{align}
    Then take \(C=C_4-C_2\), which is \(\theta\)-independent, and we prove the statement in the theorem.

\end{proof}

\section{EXPERIMENT DETAILS} 

\subsection{Numerical implementations}

When sampling from \cref{eq: inf_sde}, we first choose a grid \(\xi=\{\xi_i\}_{i=1}^{M}\) as a \(M\)-discretized domain of \(X_t\). Such a discretization varies according to different \(X_t\). We also sample \(M\) independent 1-dimensional Brownian motions \(W_t^{(1)}, W_t^{(2)}, \dots, W_t^{(M)}\) on a discrete time partition \(\{t_n\}_{n=1}^N\), such that for any \(j\)-th Brownian path, \(\Delta W^{(j)} = W^{(j)}_{t_n}-W^{(j)}_{t_{n-1}} \sim \mathcal{N}(0, \delta t)\). We sample from a system of finite-dimensional SDEs:
\begin{equation}    \label{eq: simulate_sde}
    \d X_t[\xi_i] = f(t, X_t[\xi_i])\d t + \sum^M_{j=1}[g(t, X_t[\xi_i])]_j\d W^{(j)}_t,\quad i=1,2,\dots, M
\end{equation}
\cref{eq: simulate_sde} can be used as the approximation of the original SDE in \cref{eq: inf_sde}, and can be solved using different numerical approaches. We choose Euler-Maruyama, which updates \(X_{t}[\xi_i]\) according to:
\begin{equation}
    X_{t_{n}}[\xi_i] = X_{t_{n-1}}[\xi_i] + f(t, X_{t_{n-1}}[\xi_i])\delta t + \sum^M_{j=1}[g(t, X_{t_{n-1}}[\xi_i])]_j \Delta W^{(j)},\quad X_{t_{0}}[\xi_i] = X_0[\xi_i].
\end{equation}

The objective function in \cref{eq: final_loss} can be then be approximated by:
\begin{subequations} \label{eq: hat_loss}
    \begin{align}
        \hat{L}(\theta) &= \frac{1}{2B}\sum^{N}_{n=1}\sum^{B}_{b=1}\sum^{M}_{i=1}\left[\lambda_i\left(\mathcal{G}^{(\theta)}(t_{n}, X^{(b)}_{t_n}[\xi_i]) + (\Delta t)^{-1}\cdot\{X^{(b)}_{t_n}[\xi_i] - X_{t_{n-1}}^{(b)}[\xi_i] - \Delta t \cdot f(t_{n-1}, X^{(b)}_{t_{n-1}}[\xi_i]\}\right)^2\right]\cdot \Delta t \\
        &= \frac{1}{2B}\sum^{N}_{n=1}\sum^{B}_{b=1}\sum^{M}_{i=1}\left[\lambda_i\left(\mathcal{G}^{(\theta)}(t_{n}, X^{(b)}_{t_n}[\xi_i]) + (\Delta t)^{-1}\cdot \sum^M_{j=1}[g(t_{n-1}, X^{(b)}_{t_{n-1}}[\xi_i])]_j\Delta W_j\right)^2\right]\cdot \Delta t 
    \end{align}
\end{subequations}    
where \(X^{(b)}_{t_n}[\xi_i]\) is a sample of \(X_{t_n}[\xi_i]\) indexed by \(b\), and \(\Delta t = t_{n}-t_{n-1}\). We show the training and inference algorithms in \cref{alg: training} and \cref{alg: inference}. 

All the experiments conducted in the main paper and supplementary materials are done on the platform with one NVIDIA RTX A6000 GPU and Intel(R) Xeon(R) Gold 5420+ @ 2.0GHz with 256GB memory and Ubuntu 22.04.4 LTS OS. The code for reproducing the results is available on \href{https://github.com/bookdiver/scoreoperato}{Github}.

\begin{algorithm}[ht]
\caption{Training}
\label{alg: training}
\begin{algorithmic}[1] 
\Require Domain discretization \(\xi\), initial state \(X_0[\xi]\)
\Require Total time length \(T\), time step \(\Delta t\)
\Require Batch size \(B\)
\Require Initialized \(\mathcal{G}_\theta\)
\Require Drift \(f\), diffusion \(g\)
    \State Form discrete time grid \(\{t_0=0, t_1,\dots, t_{N-1}, t_N=T\}, n=\lfloor T/\Delta t \rfloor\)
    \While{Not converged} 
        \For \(\;n \gets 1,\dots,N\)
            \For  \(\;b \gets 1,\dots,B\)
                \State Sample \(\varepsilon^{(b)}_{t_{n}}\sim\mathcal{N}(0, \Delta t\cdot\mathbf{I}_M)\), i.i.d
                \State Update \(X_t[\xi]\): \(X^{(b)}_{t_{n}}[\xi_i] \gets X^{(b)}_{t_{n-1}} + f(t_{n-1}, X^{(b)}_{t_{n-1}}[\xi_i])\Delta t + \sum^M_{j=1}[g(t_{n-1}, X^{(b)}_{t_{n-1}}[\xi_i])]_j[\varepsilon^{(b)}_{t_{n}}]_j\)
                \State Compute \(a(t_n, X_{t_n}[\xi_i])\) and extract the diagonal entries \(\lambda_i\).
            \EndFor
        \EndFor
        \State Compute \(\hat{L}(\theta)\) by \cref{eq: hat_loss}
        \State Perform gradient descent step on \(\hat{L}(\theta)\)
    \EndWhile
\end{algorithmic}
\end{algorithm}

\begin{algorithm}[ht]
\caption{Inference}
\label{alg: inference}
\begin{algorithmic}[1] 
\Require Domain discretization \(\xi'\), initial state \(Y_0[\xi']\)
\Require Total time length \(T\), time step \(\Delta t\)
\Require Pretrained \(\mathcal{G}^{(\theta)}\)
\Require Drift \(f\), diffusion \(g\)
    \State Form discrete time grid \(\{t_0=0, t_1,\dots, t_{N-1}, t_N=T\}, N=\lfloor T/\Delta t \rfloor\)
        \For \(\;n \gets 1,\dots,N\)
            \State Sample \(\varepsilon_{t_n}\sim\mathcal{N}(0, \Delta t\cdot\mathbf{I}_M)\), i.i.d
            \State Update \(\Ys_t[\xi']\): \(
                \Ys_{t_{n}}[\xi'_i] \gets \Ys_{t_{n-1}}[\xi'_i] + \{f(t_{n-1}, \Ys_{t_{n-1}}[\xi'_i]) + \mathcal{G}^{(\theta)}(t_{n-1}, \Ys_{t_{n-1}}[\xi'_i])\}\Delta t\) \\ 
                \hspace{4.5cm}\( + \sum_{j=1}^M \left\{\nabla_j[(gg^{\intercal})(t_{n-1}, \Ys_{t_{n-1}}[\xi'_i])]_j\Delta t + [g(t_{n-1}, \Ys_{t_{n-1}}[\xi'_i])]_j [\varepsilon_{t_n}]_j \right\}\)
            
        \EndFor
    \State \(\Xs_{t_n}[\xi'] \gets \Ys_{t_{N-n}}[\xi']\)
\end{algorithmic}
\end{algorithm}

\subsection{Additional experiment results} \label{sec: app_experiment_results}

\subsubsection{Quadratic functions}
\begin{figure}[ht]
    \centering
    \includegraphics[width=\textwidth]{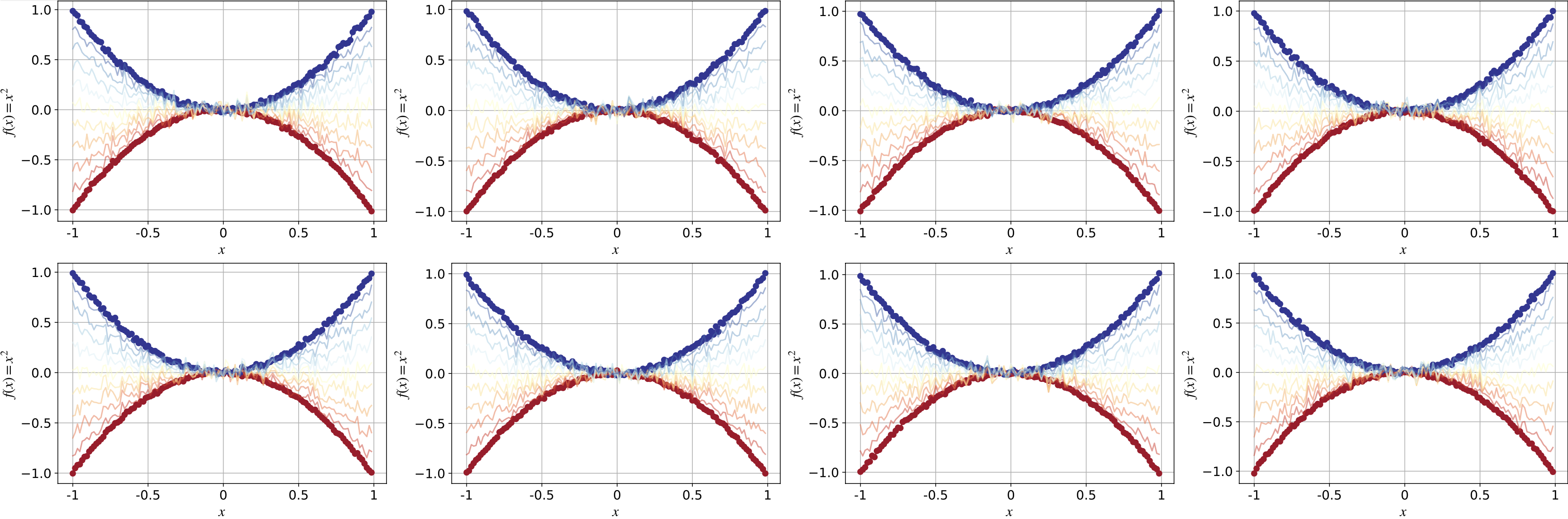}
    \caption{More samples of estimated Brownian bridges between two quadratic functions}
    \label{fig: ap1}
\end{figure}

We model the stochastic dynamics in the function space as the Brownian motion:
\begin{equation}    \label{eq: brownian}
    \d X_t = \sigma \d W_t
\end{equation}
with \(\sigma = 0.1\), which has a closed-form time reversed bridge:
\begin{equation}
    \d \Ys_t = \frac{x_0 - \Ys_t}{t}\d t + \sigma \d W_t.
\end{equation}
We designed our operator to have 6 Fourier layers. The details are shown in \cref{tab: quadratic}. We train with 10,000 i.i.d. batched samples and with a batch size of 16. We chose the Adam optimizer with an initial learning rate of 0.001 and cosine decay to 1e-5 as 80\% of the training is finished. \cref{fig: ap1} provides more samples of Brownian bridges between two quadratic functions.
\begin{table}[ht]
    \centering
    \begin{tabular}{cccccc}
       \toprule
       Layer & Input & Output & Grid size  & Fourier modes & Activation \\
       \midrule
       Lifting & \(u:\mathbb{R}\to\mathbb{R}\) & \(v_0:\mathbb{R}\to\mathbb{R}^{16}\) & 8 & - & - \\
       Down1 & \(v_0:\mathbb{R}\to\mathbb{R}^{16}\) & \(v_1:\mathbb{R}\to\mathbb{R}^{16}\) & 8 & 6 & GeLU \\
       Down2 & \(v_1:\mathbb{R}\to\mathbb{R}^{16}\) & \(v_2:\mathbb{R}\to\mathbb{R}^{32}\) & 4 & 4 & GeLU \\
       Down3 & \(v_2:\mathbb{R}\to\mathbb{R}^{32}\) & \(v_3:\mathbb{R}\to\mathbb{R}^{64}\) & 2 & 2 & GeLU \\
       Up1 & \(v_3:\mathbb{R}\to\mathbb{R}^{64}\) & \(v_4:\mathbb{R}\to\mathbb{R}^{32}\) & 2 & 2 & GeLU \\
       Up2 & \(v_4:\mathbb{R}\to\mathbb{R}^{32}\) & \(v_5:\mathbb{R}\to\mathbb{R}^{16}\) & 4 & 4 & GeLU \\
       Up3 & \(v_5:\mathbb{R}\to\mathbb{R}^{16}\) & \(v_6:\mathbb{R}\to\mathbb{R}^{16}\) & 8 & 6 & GeLU \\
       Projection & \(v_6:\mathbb{R}\to\mathbb{R}^{16}\) & \(v:\mathbb{R}\to\mathbb{R}\) & 8 & - & - \\
       \bottomrule
    \end{tabular}
    \caption{Neural operator structure for quadratic function experiments}
    \label{tab: quadratic}
\end{table}

\subsubsection{Ellipses}
\begin{figure}[ht]
    \centering
    \includegraphics[width=\textwidth]{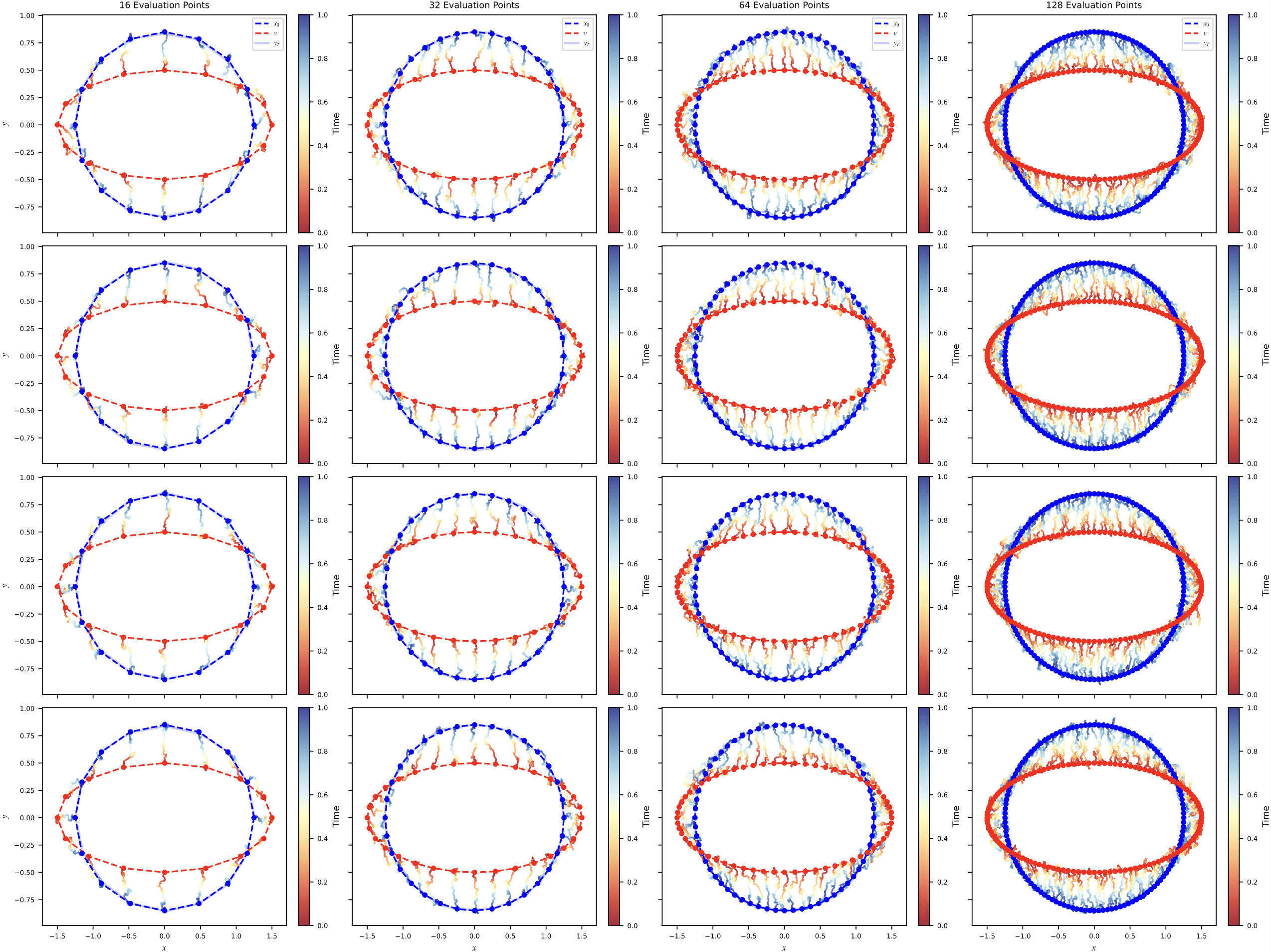}
    \caption{More samples of estimated Brownian bridges between two ellipses, evaluated under different levels of discretization}
    \label{fig: ap2}
\end{figure}

The SDE we used for modeling the stochastic development of ellipses follows \cref{eq: brownian} as well. The starting ellipse (blue) with parameters \(a=1.25, b=0.85\), and the target ellipse with parameters \(a=1.5, b=0.5\), where \(a, b\) stand for the lengths of semi-major and semi-minor axes of the ellipse. For the neural operator design, it is shown in Table \ref{tab: ellipse}. We train with 10,000 i.i.d. batched samples and with a batch size of 16. We chose the Adam optimizer with an initial learning rate of 5e-4 and cosine decay to 5e-6 as 80\% of the training is finished. We also compare the performance of our proposed neural operator architecture against the MLP structures used in \cite{baker2024conditioning}. 
Mainly, we compare the MSE between the estimated score and the true score under different levels of discretizations as shown in \cref{tab: benchmark}, while keeping the number of trained bases fixed. In \cite{baker2024conditioning}, such a hyperparameter is directly reflected on the input dimensions of the network, since the Fourier transformation has been done before the neural network, while in our implementation, it depends on the kept number of Fourier modes in the first Fourier layer after the lifting layer, because all the rest of the Fourier layers are operating on the outputs from the first layer, which determines how much information is fed into the network. All the benchmarks are conducted on the same platform with the same \textit{FLAX} \citep{flax2020github} framework to avoid potential unfairness. The code implementation of the method in \cite{baker2024conditioning} is slightly modified from the public repository \href{https://github.com/libbylbaker/infsdebridge}{https://github.com/libbylbaker/infsdebridge}. We note that our proposed neural operator structures have much lower error with smaller model sizes. We use the same Adam optimizer with the same learning rate of 2e-4 and cosine learning rate decay scheduler. The batch sizes for both models are set to be 32, and the models are both trained for 100,000 iterations. \cref{fig: ap2} shows more samples of Brownian bridges.
\begin{table}[ht]
    \centering
    \begin{tabular}{cccccc}
       \toprule
       Layer & Input & Output & Grid size  & Fourier modes & Activation \\
       \midrule
       Lifting & \(u:\mathbb{R}\to\mathbb{R}^2\) & \(v_0:\mathbb{R}\to\mathbb{R}^{16}\) & 16 & - & - \\
       Down1 & \(v_0:\mathbb{R}\to\mathbb{R}^{16}\) & \(v_1:\mathbb{R}\to\mathbb{R}^{16}\) & 16 & 8 & GeLU \\
       Down2 & \(v_1:\mathbb{R}\to\mathbb{R}^{16}\) & \(v_2:\mathbb{R}\to\mathbb{R}^{32}\) & 8 & 6 & GeLU \\
       Down3 & \(v_2:\mathbb{R}\to\mathbb{R}^{32}\) & \(v_3:\mathbb{R}\to\mathbb{R}^{64}\) & 4 & 4 & GeLU \\
       Up1 & \(v_3:\mathbb{R}\to\mathbb{R}^{64}\) & \(v_4:\mathbb{R}\to\mathbb{R}^{32}\) & 4 & 4 & GeLU \\
       Up2 & \(v_4:\mathbb{R}\to\mathbb{R}^{32}\) & \(v_5:\mathbb{R}\to\mathbb{R}^{16}\) & 8 & 6 & GeLU \\
       Up3 & \(v_5:\mathbb{R}\to\mathbb{R}^{16}\) & \(v_6:\mathbb{R}\to\mathbb{R}^{16}\) & 16 & 8 & GeLU \\
       Projection & \(v_6:\mathbb{R}\to\mathbb{R}^{16}\) & \(v:\mathbb{R}\to\mathbb{R}^2\) & 16 & - & - \\
       \bottomrule
    \end{tabular}
    \caption{Neural operator structure for ellipse experiments}
    \label{tab: ellipse}
\end{table}

\vfill

\subsubsection{Spheres}
\begin{figure}[ht]
    \centering
    \includegraphics[width=\textwidth]{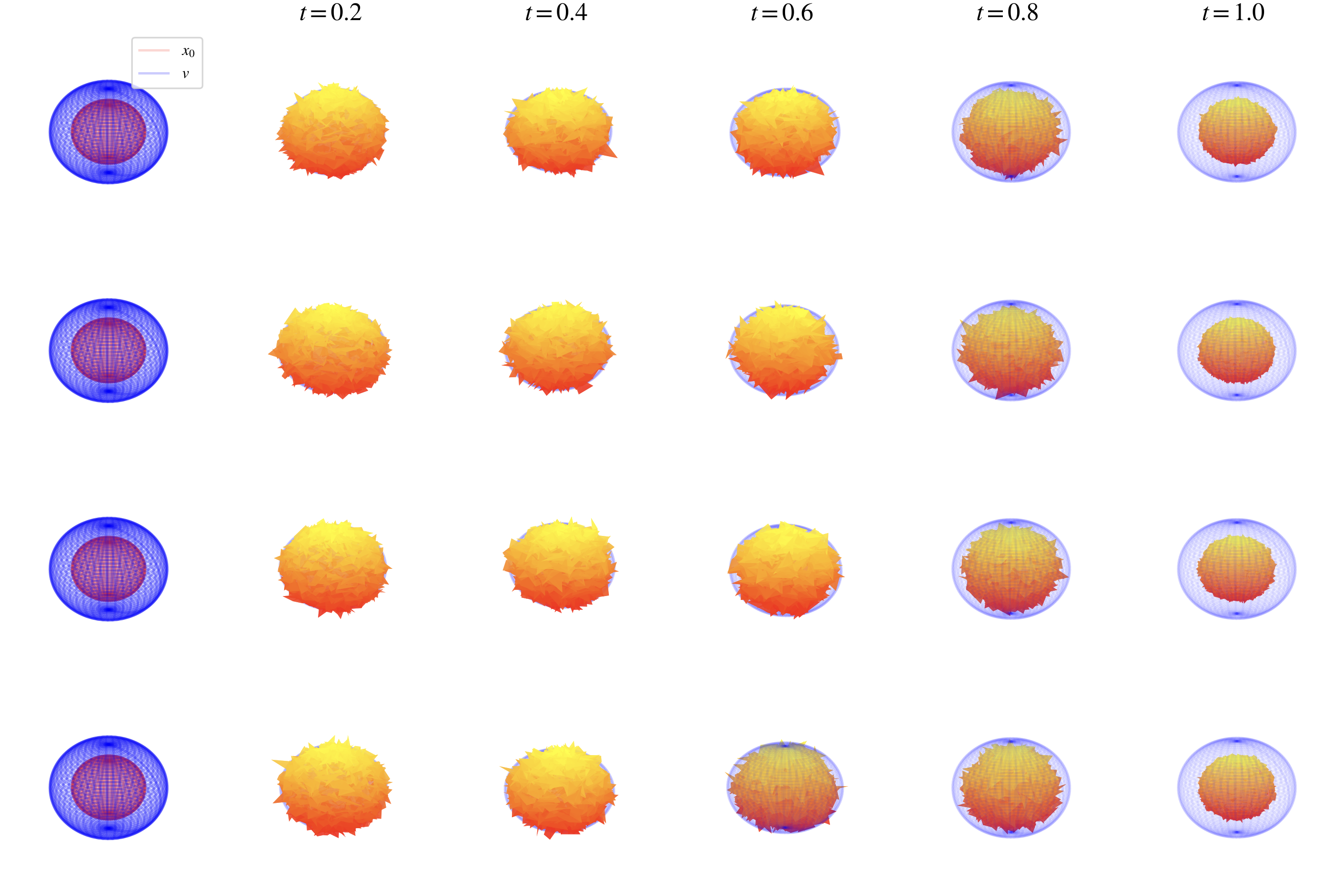}
    \caption{More samples of estimated Brownian bridges between two spheres}
    \label{fig: ap3}
\end{figure}

Unlike the previous two experiments, spheres should be treated as functions \(\R^2 \to (\mathcal{S}^2\subset\R^3)\). The samples then have the shape of \((m, m, 3)\) and we modify the neural operator structure to fit in it. \cref{tab: ellipsoid} shows the detailed structure and \cref{fig: ap3} shows more samples.

\begin{table}[ht]
    \centering
    \begin{tabular}{cccccc}
       \toprule
       Layer & Input & Output & Grid size  & Fourier modes & Activation \\
       \midrule
       Lifting & \(u:\mathbb{R}^2\to\mathbb{R}^3\) & \(v_0:\mathbb{R}^2\to\mathbb{R}^{32}\) & (16, 16) & - & - \\
       Down1 & \(v_0:\mathbb{R}^2\to\mathbb{R}^{32}\) & \(v_1:\mathbb{R}^2\to\mathbb{R}^{32}\) & (16, 16) & (12, 12) & GeLU \\
       Down2 & \(v_1:\mathbb{R}^2\to\mathbb{R}^{32}\) & \(v_2:\mathbb{R}^2\to\mathbb{R}^{64}\) & (16, 16) & (8, 8) & GeLU \\
       Down3 & \(v_2:\mathbb{R}^2\to\mathbb{R}^{64}\) & \(v_3:\mathbb{R}^2\to\mathbb{R}^{128}\) & (8, 8) & (4, 4) & GeLU \\
       Up1 & \(v_3:\mathbb{R}^2\to\mathbb{R}^{128}\) & \(v_4:\mathbb{R}^2\to\mathbb{R}^{64}\) & (8, 8) & (4, 4) & GeLU \\
       Up2 & \(v_4:\mathbb{R}^2\to\mathbb{R}^{64}\) & \(v_5:\mathbb{R}^2\to\mathbb{R}^{32}\) & (16, 16) & (8, 8) & GeLU \\
       Up3 & \(v_5:\mathbb{R}^2\to\mathbb{R}^{64}\) & \(v_6:\mathbb{R}^2\to\mathbb{R}^{32}\) & (16, 16) & (12, 12) & GeLU \\
       Projection & \(v_6:\mathbb{R}^2\to\mathbb{R}^{32}\) & \(v:\mathbb{R}^2\to\mathbb{R}^3\) & (16, 16) & - & - \\
       \bottomrule
    \end{tabular}
    \caption{Neural operator structure for sphere experiments}
    \label{tab: ellipsoid}
\end{table}

\subsubsection{Butterflies} \label{sec: app_butterflies_experiments}

\begin{figure}[ht]
    \centering
    \includegraphics[width=\textwidth]{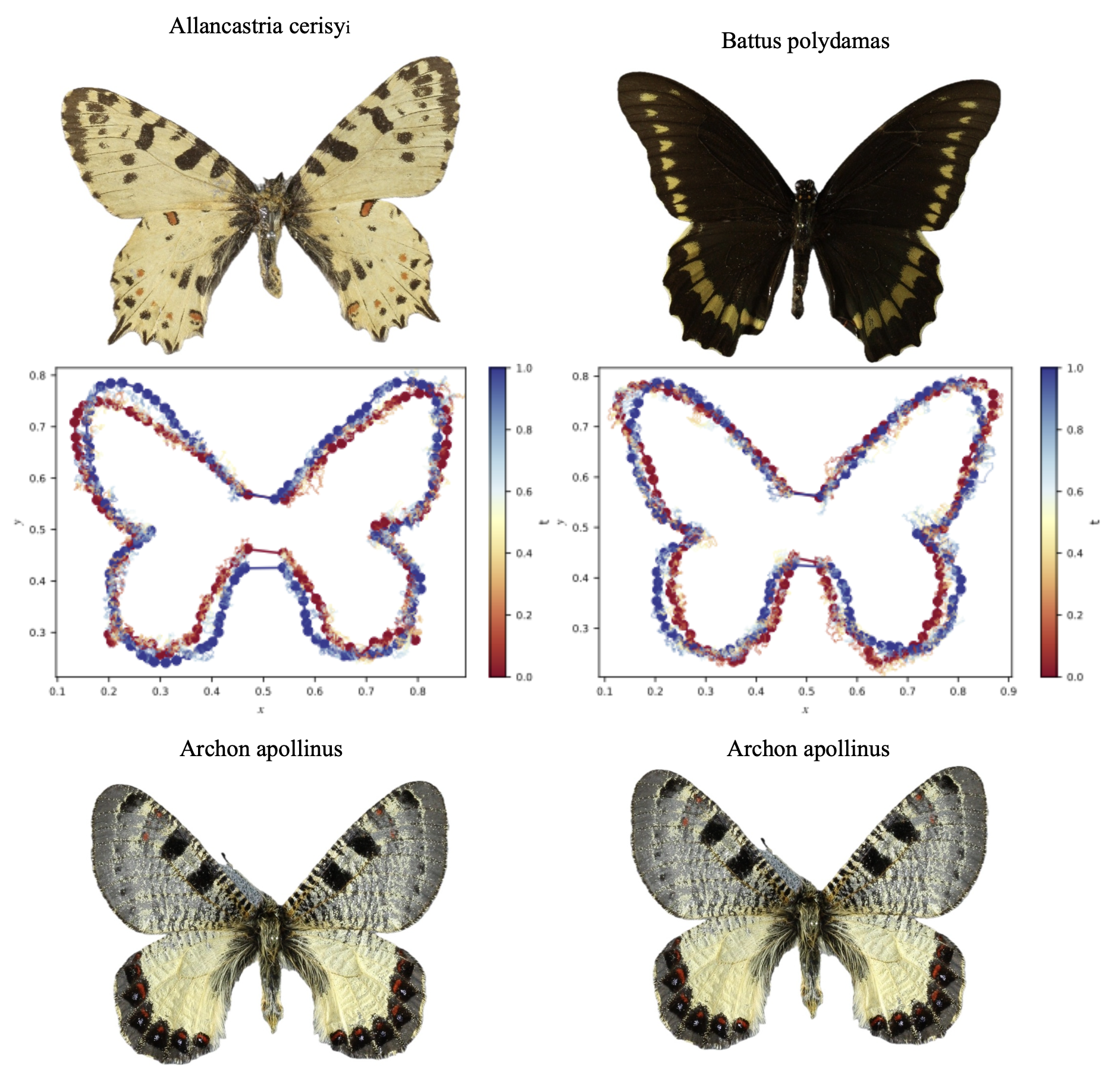}
    \caption{The additional bridge simulations between different species, the bridges are constructed between \textit{Archon apollinus} (blue) and \textit{Allancastria cerisyi/Battus polydamas} (red).}
    \label{fig: ap4}
\end{figure}

\begin{figure}[ht]
    \centering
    \includegraphics[width=\textwidth]{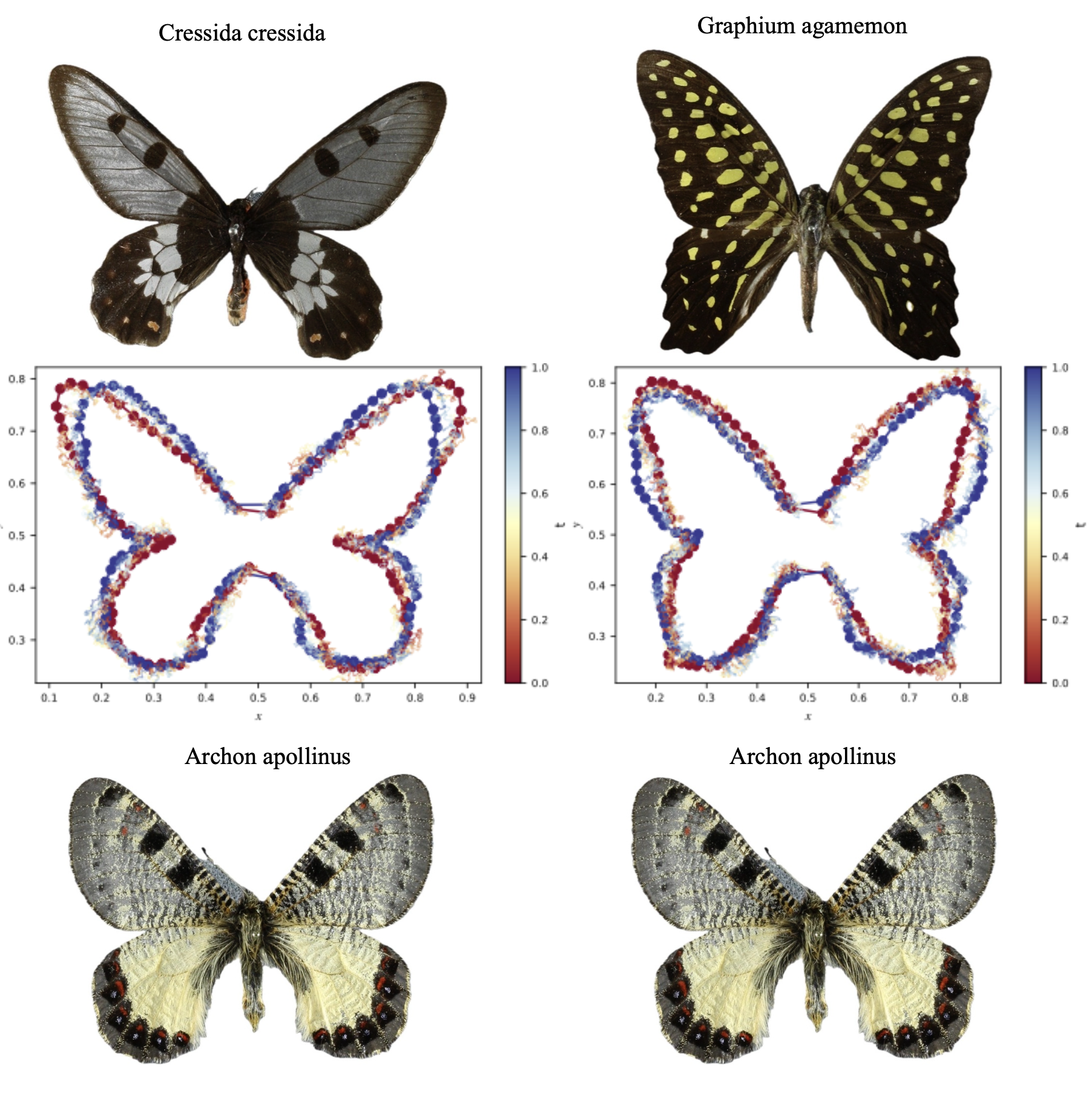}
    \caption{The additional bridge simulations between different species, the bridges are constructed between \textit{Archon apollinus} (blue) and \textit{Cressida cressida/Graphium agamemon} (red). }
    \label{fig: ap5}
\end{figure}

As introduced in \cref{eq: kunita_sde}, we let the butterfly shapes be subsets of \(\mathbb{R}^2\), and choose the kernel \(k\) to be the 2D-Gaussian kernel, i.e., \(k(x,y) = \sigma\exp(-\|x-y\|^2/\kappa)\). Specifically, we set the domain in \(\mathbb{R}^2\) to be finite as \([-0.5, 1.5]^2\) and discretize it with the resolution of \(50\times 50\). We then choose the kernel with \(\sigma=0.04\) and \(\kappa=0.02\), which allows correlations only within small areas. We then design the operator as \cref{tab: butterfly}. We train with 20,000 i.i.d. batched samples and with a batch size of 16. We chose the Adam optimizer with an initial learning rate of 0.001 and cosine decay to 1e-5 as 80\% of the training is finished. \cref{fig: ap4} and \cref{fig: ap5} show the nonlinear shape bridges between five chosen butterfly species and their specimen photos.

\begin{table}[ht]
    \centering
    \begin{tabular}{cccccc}
       \toprule
       Layer & Input & Output & Grid size  & Fourier modes & Activation \\
       \midrule
       Lifting & \(u:\mathbb{R}\to\mathbb{R}^2\) & \(v_0:\mathbb{R}\to\mathbb{R}^{16}\) & 32 & - & - \\
       Down1 & \(v_0:\mathbb{R}\to\mathbb{R}^{16}\) & \(v_1:\mathbb{R}\to\mathbb{R}^{16}\) & 32 & 16 & GeLU \\
       Down2 & \(v_1:\mathbb{R}\to\mathbb{R}^{16}\) & \(v_2:\mathbb{R}\to\mathbb{R}^{32}\) & 16 & 8 & GeLU \\
       Down3 & \(v_2:\mathbb{R}\to\mathbb{R}^{32}\) & \(v_3:\mathbb{R}\to\mathbb{R}^{64}\) & 8 & 6 & GeLU \\
       Down4 & \(v_3:\mathbb{R}\to\mathbb{R}^{64}\) & \(v_4:\mathbb{R}\to\mathbb{R}^{64}\) & 8 & 6 & GeLU \\
       Up1 & \(v_4:\mathbb{R}\to\mathbb{R}^{64}\) & \(v_5:\mathbb{R}\to\mathbb{R}^{64}\) & 8 & 6 & GeLU \\
       Up2 & \(v_5:\mathbb{R}\to\mathbb{R}^{64}\) & \(v_6:\mathbb{R}\to\mathbb{R}^{32}\) & 8 & 6 & GeLU \\
       Up3 & \(v_6:\mathbb{R}\to\mathbb{R}^{32}\) & \(v_7:\mathbb{R}\to\mathbb{R}^{16}\) & 16 & 8 & GeLU \\
       Up4 & \(v_7:\mathbb{R}\to\mathbb{R}^{16}\) & \(v_8:\mathbb{R}\to\mathbb{R}^{16}\) & 32 & 16 & GeLU \\
       Projection & \(v_8:\mathbb{R}\to\mathbb{R}^{16}\) & \(v:\mathbb{R}\to\mathbb{R}^2\) & 32 & - & - \\
       \bottomrule
    \end{tabular}
    \caption{Neural operator structure for butterfly experiments}
    \label{tab: butterfly}
\end{table}

\end{document}